\documentclass[10pt,twocolumn,letterpaper]{article}

\usepackage{iccv}              
\usepackage{url}
\usepackage{soul}
\usepackage{amsmath}
\usepackage{amssymb}
\usepackage{graphicx}
\usepackage{lipsum}
\usepackage{makecell}
\usepackage{adjustbox}
\usepackage[utf8]{inputenc} 
\usepackage[T1]{fontenc}    
\usepackage{url}            
\usepackage{booktabs}       
\usepackage{amsfonts}       
\usepackage{nicefrac}       
\usepackage{microtype}      
\usepackage{multirow}
\usepackage{float}
\usepackage{subcaption}
\usepackage{dsfont}
\usepackage{tcolorbox}
\tcbuselibrary{breakable,xparse,skins}
\usepackage{caption}
\usepackage{xspace}
\usepackage{amsmath}
\usepackage{tabularx}
\usepackage{bm}
\usepackage{mathtools}
\usepackage{wrapfig}
\usepackage{tikz}
\usepackage{ifthen}
\usetikzlibrary{
    positioning, 
    shapes,
    shapes.geometric, 
    shapes.multipart,
    calc,
    arrows,
    arrows.meta,
    fit,
    backgrounds}
\usepackage[inline]{enumitem}
\graphicspath{ {./figures/} }
\definecolor{iccvblue}{rgb}{0.21,0.49,0.74}
\usepackage[pagebackref,breaklinks,colorlinks,allcolors=iccvblue]{hyperref}

\def\paperID{10393} 
\def\confName{ICCV}
\def\confYear{2025}

\title{TAPNext: Tracking Any Point (TAP) as Next Token Prediction}

\author{Artem Zholus\thanks{Work done duing internship at Google DeepMind}\textsuperscript{  \;\rm 2,\rm 4,\rm 5} \and Carl Doersch\textsuperscript{\rm 1} \and Yi Yang\textsuperscript{\rm 1} \and Skanda Koppula\textsuperscript{\rm 1,\rm 3} \and Viorica Patraucean\textsuperscript{\rm 1} \and Xu Owen He\textsuperscript{\rm 1} \and Ignacio Rocco\textsuperscript{\rm 1} \and Mehdi S. M. Sajjadi\textsuperscript{\rm 1} \and Sarath Chandar\textsuperscript{\rm 2,\rm 4,\rm 5,\rm 6} \and Ross Goroshin\textsuperscript{\rm 1,\rm 4,\rm 7} \\[3mm]
\small{
\textsuperscript{\rm 1}Google DeepMind \;\;\;
\textsuperscript{\rm 2}Chandar Research Lab \;\;\;
\textsuperscript{\rm 3}University College London \;\;\;
\textsuperscript{\rm 4}Mila - Quebec AI Institute \;\;\;
}\\
\small{
\textsuperscript{\rm 5}Polytechnique Montreal \;\;\; \textsuperscript{\rm 6}Canada CIFAR AI Chair} \;\;\;
\textsuperscript{\rm 7}Université de Montréal \;\;\;
}

\begin{document}
\maketitle
\begin{abstract}
\label{sec:abstract}
	Tracking Any Point (TAP) in a video is a challenging computer vision problem with many demonstrated applications in robotics, video editing, and 3D reconstruction. Existing methods for TAP rely heavily on complex tracking-specific inductive biases and heuristics, limiting their generality and potential for scaling. To address these challenges, we present TAPNext, a new approach that casts TAP as sequential masked token decoding. Our model is causal, tracks in a purely online fashion, and removes tracking-specific inductive biases. This enables TAPNext to run with minimal latency, and removes the temporal windowing required by many existing state of art trackers. Despite its simplicity, TAPNext achieves a new state-of-the-art tracking performance among both online and offline trackers. Finally, we present evidence that many widely used tracking heuristics emerge naturally in TAPNext through end-to-end training. The TAPNext model and code can be found at \url{https://tap-next.github.io}.
\end{abstract}
    
\section{Introduction}
\label{sec:intro}

\begin{figure*}
    \centering
    \begin{subfigure}[b]{0.3\textwidth}
        \centering
        \includegraphics[width=\columnwidth]{figures/teaser/horse_queries.png}
        \caption{Grid of query points}
    \end{subfigure}
    \begin{subfigure}[b]{0.34\textwidth}
        \centering
        \includegraphics[width=\columnwidth]{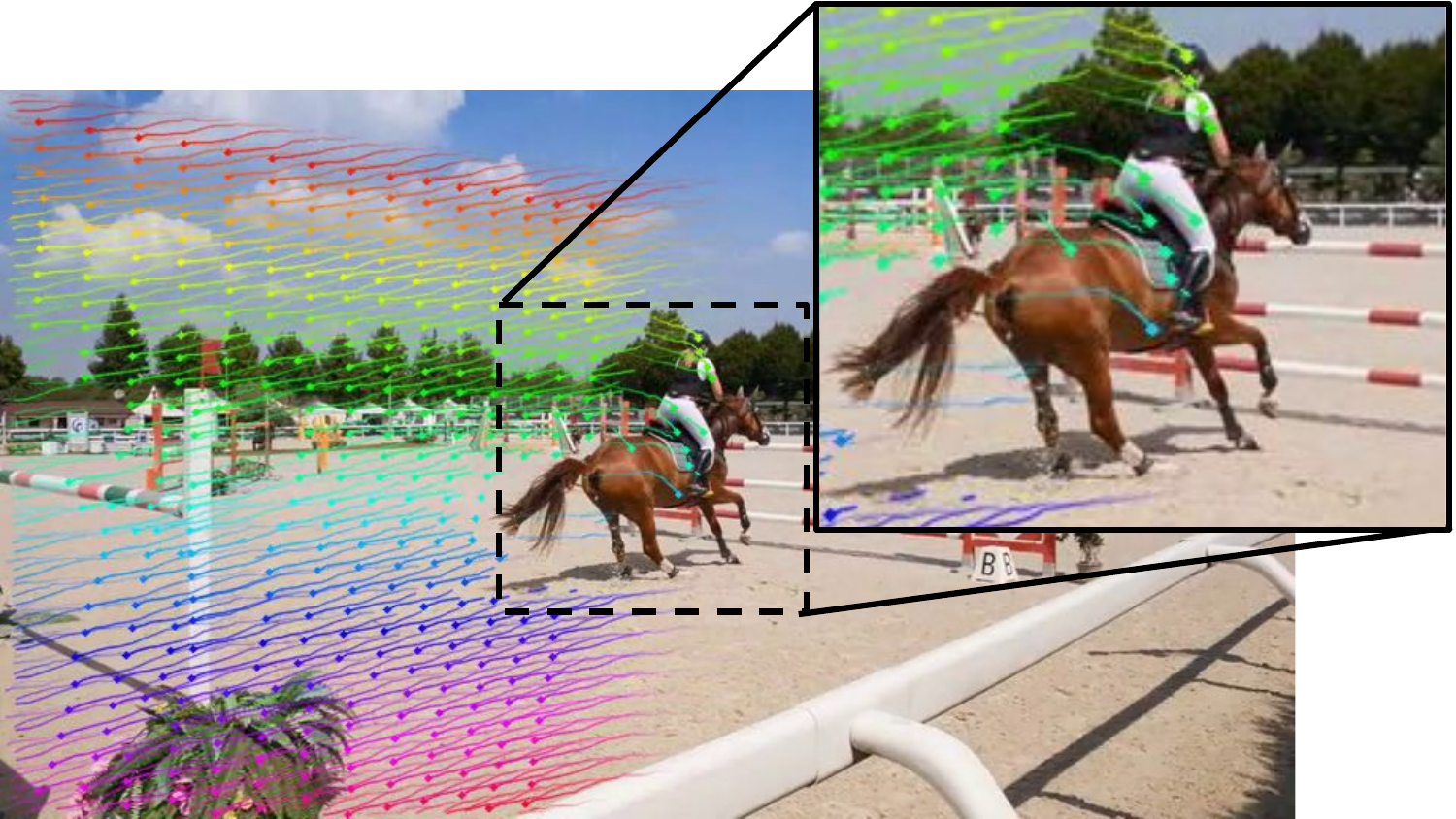}
        \caption{CoTracker3~\cite{cotracker3}}
    \end{subfigure}
    \begin{subfigure}[b]{0.34\textwidth}
        \centering
        \includegraphics[width=\columnwidth]{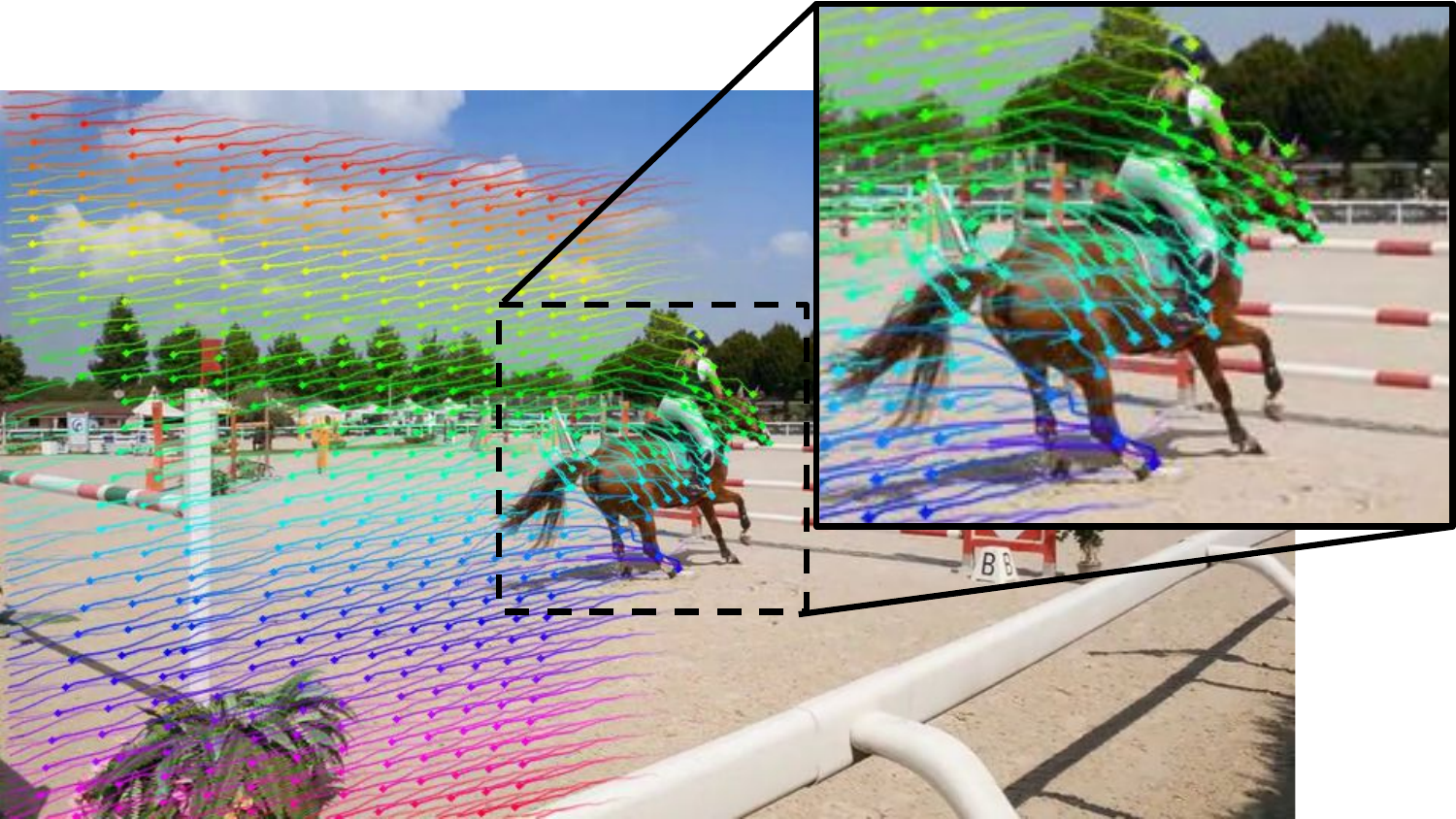}
        \caption{TAPNext}
    \end{subfigure}
    \caption{Dense grid tracking with TAPNext. We show (a) the query points on the first frame of the video, (b) the resulting tracks on the final frame of the video for  CoTracker3~\cite{cotracker3}, and (c) our proposed TAPNext method.}
    \label{fig:dense_tracking}
\end{figure*}

``\emph{Correspondence, correspondence, and correspondence!}'' was Takeo Kanade's answer when asked for the three most fundamental problems in computer vision. In early research, image-level correspondence was pivotal for understanding motion \cite{huang1994motion}, estimating depth~\cite{hartley2003multiple} or performing 3D reconstruction from photo collections~\cite{szeliski2022computer}. Today,  correspondence has experienced a resurgence as Tracking Any Point (TAP) in video: outputs are dense like optical flow, but also provide long-range correspondence as in keypoint detection. In TAP a model is tasked to track a large number of \emph{query} points, which correspond to surfaces on solid objects, over the course of a video (Figure \ref{fig:dense_tracking}). Because point tracking is such a generic task, yielding highly detailed spatiotemporal information, point trackers are potentially useful in almost any downstream computer vision application. Indeed, the three years since TAP's inception it has already been applied to robotics~\cite{vecerik2024robotap,abbeel,bharadhwaj2024track2act,yuan2024general,bharadhwaj2024gen2act,li2024unifying}, action recognition~\cite{kumar2025trajectory}, 3D (dynamic) reconstruction~\cite{wang2023omnimotion,wang2023visual,xiao2024spatialtracker}, (controllable) video generation and editing~\cite{tapir,yu2023videodoodles,wu2025draganything,li2024image,zhou2024trackgo}, zoology~\cite{polajnar2024wing}, and medicine~\cite{schmidt2024surgical}.

Real-world training data for TAP is scarce~\citep{tapvid, patraucean2023perception, vecerik2024robotap, balasingam2024drivetrack,koppula2024tapvid}, so prior works have trained models mainly on synthetic data ~\cite{greff2022kubric,zheng2023pointodyssey}.  These works have argued that complex inductive biases and custom architectures are required to bridge the sim-to-real gap.  However, making architectures \textit{too} restrictive can limit scalability.  In this work, we ask what architectural (e.g. cost-volume) and algorithmic (e.g. iterative refinement) design decisions can be relaxed in order to build a more scalable, simpler, yet performant model. Our key finding is that only two ingredients are necessary to achieve good TAP performance: 1) a shared, spatio-temporal bank of input tokens, where some tokens encode image patches and others encode trajectories, all processed jointly by a unified model, and 2) a \textit{simple} recurrence across time, in the form of a real-valued State-Space Model (SSM)~\cite{gemmateam2024gemmaopenmodelsbased}, and 3) an output encoding that represents distributions instead of point estimates.  Specifically, given our input encoding, we find that \textit{TRecViT}~\cite{patraucean2024trecvitrecurrentvideotransformer}--which simply interleaves (spatial) transformer layers and (temporal) SSM layers can be used off-the-shelf to predict point tracks.  In fact, when fine-tuned on real data using a previously published procedure~\cite{bootstap}, this architecture achieves state-of-the-art performance without any complex post-processing or iterative-inference. This stands in stark contrast to all other SoTA TAP methods which use windowed inference, iterative refinement, and temporal smoothness priors. The architecture consists of interleaved layers of 1) temporally-recurrent computations processing ``temporal tubes'' \cite{kwak2015unsupervised} which are sequences of individual image and point tokens, and 2) transformer layers which process spatial information by attending across image \& point tokens, per-frame.

Our contributions include:
\begin{enumerate}
    \item We propose TAPNext, a new point tracking model, which performs point tracking via imputation of unknown tokens in a masked decoding fashion. TAPNext achieves state-of-the-art online tracking performance using open source architectural components (SSM \& ViT) in a backbone first proposed by \citep{patraucean2024trecvitrecurrentvideotransformer}, without using any tracking-specific inductive biases.
  \item We show that while TAPNext does not include iterative or windowed inference, test-time optimization, cost volumes, feature interpolation, token feature engineering, or local search windows, 
  some of these heuristics naturally emerge in TAPNext from end-to-end training on a large synthetic dataset.
  \item We demonstrate that recurrent networks (specifically SSM layers) can track points causally online, in a per-frame manner, generalizing to much longer videos compared to those seen in training. Linear recurrence of SSMs allows temporal processing to be parallelized in the offline setting \citep{s4}. 
  \item We are immediately open sourcing the inference code and model weights of TAPNext and BootsTAPNext. We plan to open source our training code base in the near future. 
\end{enumerate}
\section{Background}

\subsection{Optical Flow and Point Tracking}



Frame-to-frame correspondence, or optical flow, has many modern deep learning approaches that solve the task, like FlowNet \cite{dosovitskiy2015flownet} or RAFT \cite{teed2020raft}, which improve on classical methods like Lucas-Kanade \cite{baker2004lucas}.
Unfortunately, optical flow based correspondence models incur significant drift over long time horizons and traditionally do not address occlusion. Long term point tracking, or Tracking any Point (TAP), was introduced to bridge this gap \citep{tapvid}. 

Prior TAP works often formulate tracking as a two step process: first, per-frame encoding and matching via cost-volume computation, followed by track refinement \citep{pips, tapvid, tapir, cotracker}.
The cost-volume computes the inner product between a query point feature and all candidate matching features in a video (albeit often at a lower spatial resolution). Cost volumes are computed independently for every query point, and introduce an inductive bias that casts tracking as a per-frame appearance matching problem in feature space. Motion information is integrated on top of this computation as an iterative refinement step \citep{pips, tapir} or as an attention operation that processes the cost volume correlation features \citep{cotracker, li2024taptr}. Such architectures involve many heuristic design elements, including the differentiable $argmax$ operation \cite{tapir}, bi-linear interpolation of features \cite{pips}, restricted search windows \cite{cho2024local}, and windowed inference \cite{cotracker}. \emph{Our aim is to develop a conceptually simple architecture without imposing these strong inductive biases on exactly how the tracking problem should be solved, and without explicit per-frame appearance matching.}

Many trackers also rely on using future frames to produce outputs for the current frame, limiting their applicability in real-time scenarios. For this reason, online point tracking has gained increasing attention, particularly in robotics \cite{vecerik2024robotap}. Although several local window-based inference methods claim to offer online tracking, their reliance on large temporal windowing and the transfer of points only between consecutive windows often leads to tracking failures, especially during long-term occlusions in the middle of a video \cite{pips, cotracker}. As a result, these approaches tend to perform poorly on long-term tracking benchmarks \cite{patraucean2023perception}. At the same time, purely online, per-frame point tracking \cite{trackon,bootstap,vecerik2024robotap} has emerged with many applications in robotics \cite{abbeel,vecerik2024robotap} and generative modeling \cite{zhou2024trackgo}.

In this work we show that these limitations can be addressed through the use of recurrent state architectures (e.g. SSMs). The recurrent state allows the model to maintain temporal coherence, even during long-term occlusions, by effectively capturing and preserving the dynamics of the tracked points across time.



\begin{figure*}[t!]
    \centering
    \resizebox{\textwidth}{!}{
        \input{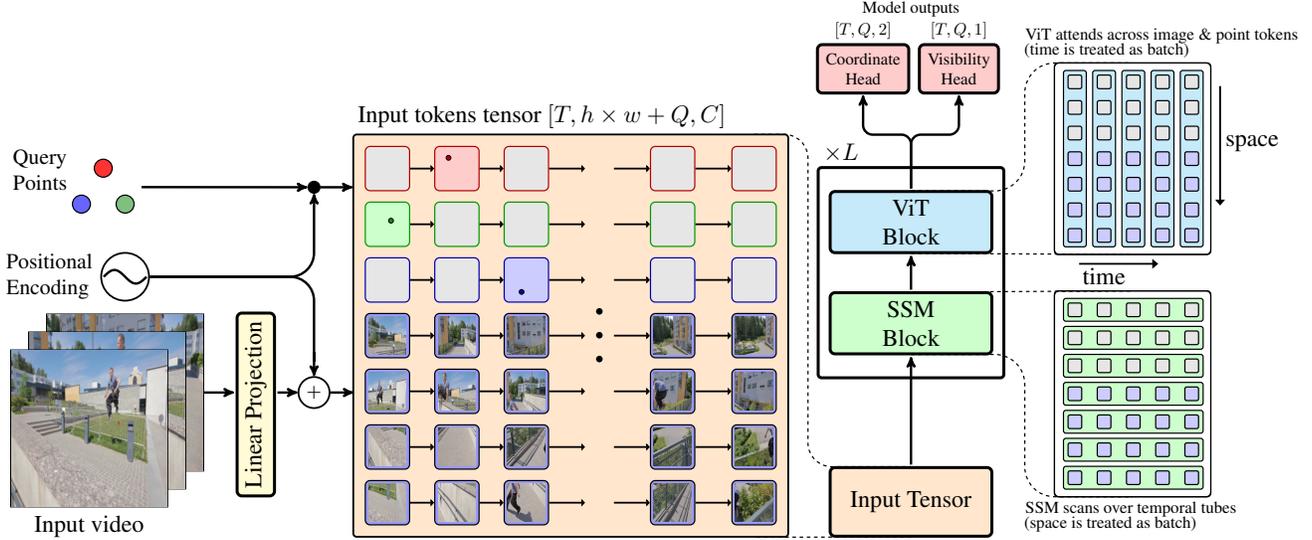}
    }
    \caption{TAPNext performs tracking via imputation of unknown point coordinates given known ones (query points and the video). This imputation happens via temporal masked decoding of tokens: video tokens are concatenated with point coordinate tokens and the latter inject point query information via positional encoding. }
    \label{fig:input_scheme}
\end{figure*}

\subsection{Masked decoding and token imputation}
Self-supervised models such as Masked Autoencoders (MAE) \cite{he2022masked}, VideoMAE \cite{tong2022videomae} and MultiMAE \cite{bachmann2022multimae} have demonstrated the ability to learn rich, generalized representations of images and videos without requiring large annotated datasets. These models mask portions of the input data and task the network with reconstructing the missing parts, thereby forcing it to learn semantically meaningful representations that capture both spatial and temporal coherence. From the perspective of MultiMAE, our framework can be interpreted as treating point and image tokens as two different modalities. Similar formulations have also been used to train detection and segmentation models in the supervised setting \citep{carion2020detr, kirillov2023segment}.

\subsection{Recurrent Models}
Recurrent models have long been used in video analysis tasks to capture temporal dependencies in sequential data. The use of Recurrent Neural Networks (RNNs) \cite{rumelhart1986learning}, and more specifically LSTM networks \cite{hochreiter1997long}, has been widespread in sequence modeling tasks \cite{karpathy2015deep}, and even object tracking \cite{gordon2018re3}. These models process each frame in a sequence by maintaining a hidden state that evolves over time, allowing the model to learn temporal dynamics and propagate information from past frames \cite{van2016wavenet}. The recurrent state has the capacity to serve multiple functions when applied to tracking problems, from motion modelling to maintaining a representation of the tracked object's appearance online \citep{gordon2018re3, vecerik2024robotap}. Unlike other recently proposed state-space model (SSM) architectures for video \citep{li2024videomamba} which are applied spatially (e.g. across image patches/tokens), we apply our SSM \citep{botev2024recurrentgemma} only on temporal tubes and use the ViT layers for spatial processing \citep{patraucean2024trecvitrecurrentvideotransformer}. Our work is the first to apply SSMs to the TAP task. 


\section{TAPNext}
\label{label:method}

TAPNext processes videos of $T$ frames, where each frame is an RGB image of size $H\times W$ pixels. Like other trackers, TAPNext receives $Q$ query points, each consisting of $(t, x, y)$ denoting time ($t$) and $x,y$ coordinates of the query point in the video. We can consider RGB frames and query points as two different input \emph{modalities} that are jointly processed, as illustrated in Figure ~\ref{fig:input_scheme}. For the visual modality, like other ViT-based methods, we partition each image in the video into $h\times w$ non-overlapping image patches. These image patches are linearly projected to a $C$ dimensional space and spatial positional embeddings are added to them. The resulting tensor has a shape of $[T, h \times w, C]$ which we refer to as the (input) \emph{image tokens}. Next, we detail how the $Q$ query points are encoded into tokens and how the tracking task is formulated as a masked decoding problem.

\textbf{Tracking as Masked Decoding.} TAPNext converts each query into a sequence of tokens, one per frame. For the frame which corresponds to the query, this token is initialized with the query coordinates, the point tokens corresponding to the remaining frames are initialized with a \textit{mask token}; thus, inference is done by filling in the mask tokens.
We refer to mask and query tokens collectively as point track tokens. In more detail, assuming we have \emph{one} query point per track, of the form $(t, x, y)$, we create $T$ point track tokens for each of the $Q$ tracks. For each track, the token corresponding to the query point located at time $t$ is filled with the positional embedding values corresponding to the query coordinates, $(x, y)$. The remaining $Q \times (T-1)$ point tokens are filled with the same learned \textit{mask token} value. We do not apply temporal positional embeddings, as there is no self-attention across time; in fact, we find temporal position embeddings can harm generalization across sequence length (i.e. to longer videos). The tensor of tokens representing image patches (of shape $[T, h\times w, C]$) is concatenated with the tensor of point track tokens (of shape $[T, Q, C]$) along the spatial dimension, resulting in an input tensor of shape $[T, h\times w + Q, C]$, as shown in the center of Figure ~\ref{fig:input_scheme}. To predict the masked point coordinates, we simply take the $T\times Q$ point track tokens from the output of the model and pass them to prediction heads. Note that the model learns to copy the one known query to the output. This process is very similar to the \emph{decoding process} used in Masked Autoencoder (MAE) \citep{he2022masked} or VideoMAE \citep{tong2022videomae} except that unknown query tokens are considered to be ``masked'', while image tokens are always visible.  




\textbf{Separating spatial and temporal processing.} 
We aim to train a tracking architecture with minimal tracking-specific inductive biases and post-processing, relying as much as possible on general-purpose components. For this purpose we adapt TRecViT \citep{patraucean2024trecvitrecurrentvideotransformer}, which consists of two building blocks: SSM blocks and ViT blocks. This architecture processes the aforementioned input tokens ($T \times (h\times w + Q)$ in total) through $L$ layers. The RecurrentGemma\footnote{\href{https://github.com/google-deepmind/recurrentgemma}{https://github.com/google-deepmind/recurrentgemma}} \citep{griffin,botev2024recurrentgemma} SSM block performs temporal processing via linear recurrence followed by a ViT block\footnote{\href{https://github.com/google-research/big_vision}{https://github.com/google-research/big\_vision}} \citep{dosovitskiy2021imageworth16x16words} that performs spatial processing with attention. 



RecurrentGemma is a causal recurrent model which was initially proposed for causal language modeling. Like other SSM models, RecurrentGemma \citep{griffin,botev2024recurrentgemma} uses linear recurrence which allows it to be parallelized in time and effectively train on very long sequences.
The computational graph for every TAPNext layer is visualized on the right side of Figure \ref{fig:input_scheme}. Specifically, each TAPNext layer has an SSM block that performs linear recurrence over the temporal dimension ($T$) while treating the $h\times w + Q$ token dimension as a batch. This architecture operates on ``spatiotemporal tubes'' \citep{kwak2015unsupervised, patraucean2024trecvitrecurrentvideotransformer} in the first layer, before any mixing across spatial tokens. Because the known query points are fed into the model only at a specific timestep, the model must learn to propagate this information in the recurrent state to all subsequent time steps. These recurrent temporal blocks allow our model to successfully track points in long videos without increasing the computational complexity or resorting to windowed inference schemes \citep{tapir, pips, cotracker}. The second stage of every TAPNext layer is a standard ViT block that attends between image and point track tokens ($h\times w + Q$), treating the time dimension ($T$) as batch.
 
\begin{figure}
  \begin{center}
    \resizebox{\linewidth}{!}{
        \input{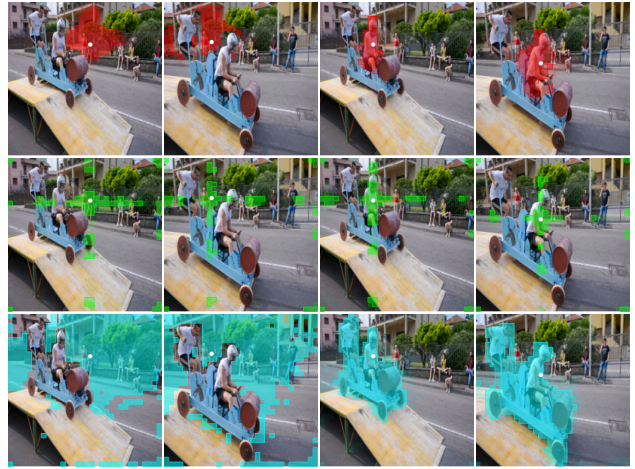}
    }
  \end{center}
  \vspace{-12pt}
  \caption{Three attention patterns learned by TAPNext. We visualize attention maps where the attention queries are the point track tokens and the keys are image tokens, which correspond to $8\times8$ patches. Each row is a certain (layer, head) pair. We observe patterns: \textbf{(top)} Cost-volume-like attention head; \textbf{(middle)} Coordinate-based readout head; \textbf{(bottom)} motion-cluster-based readout head. Note that these are just intermediate heads in the backbone. Higher resolution image and full attention maps in Appendix \ref{app:attn_vis}.}
  \label{fig:spatial_attn}
\end{figure}
\begin{figure*}
  \begin{center}
    \resizebox{\linewidth}{!}{
        \input{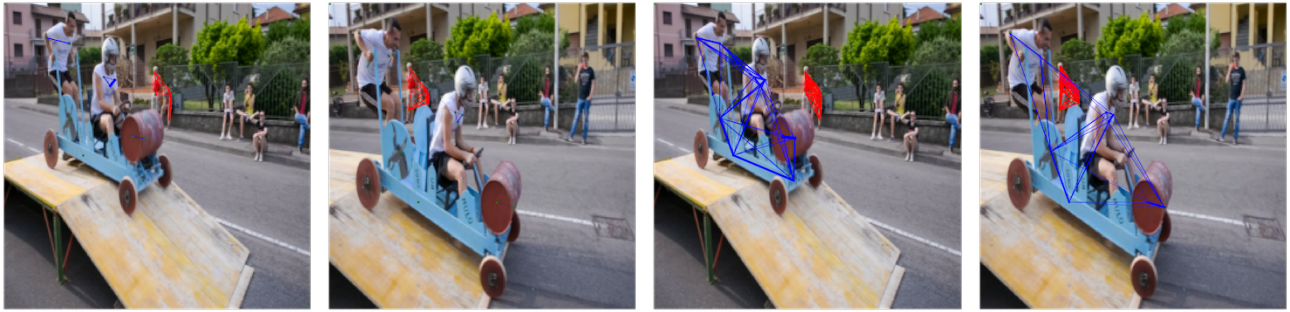}
    }
  \end{center}
  \vspace{-12pt}
  \caption{Point-to-point attention map visualizations. Tracked points are nodes and (scaled) attention weights are edges, the thicker the edge the higher the weight between points. Two frames from a video are used to visualize two attention layers. 
  Note that in all images we see strong attention between points on objects that are moving together.
  See higher resolution images in Appendix \ref{app:attn_vis}.
  }
  \label{fig:query_attn}
  \vspace{-15pt}
\end{figure*}

\textbf{Model interpretation}. 
Each ViT block in TAPNext performs full self-attention over $h\times w + Q$ tokens (in every frame). This means every track token gets updated with ($h\times w$) spatial image tokens and ($Q - 1$) other track tokens. In general each attention matrix can be subdivided into four sections: point-to-image, point-to-point, image-to-image, and image-to-point. This suggests a simple way to visualize TAPNext's ViT layers: image-to-point attention weights show how information is read from the image (Figure \ref{fig:spatial_attn}) and point-to-point attention weights show how points interact in their representations (Figure \ref{fig:query_attn}). See Section \ref{sec:interp} for a more detailed discussion.



\textbf{Prediction heads and loss functions.} 
For each timestep, each layer of our model outputs the same number of tokens as the input, i.e. $h \times w + Q$, with $Q$ query tokens. The goal of the point tracking task is to predict the coordinate and visibility for each queried point in each frame. Therefore, we simply take $T\times Q$ output track tokens  and feed them to prediction heads (MLP networks): one head predicts $(x, y) \in [0, H]\times [0, W]$ coordinates and the other predicts binary visibility. The outputs are trained using ground truth track coordinates and visibility flag at the corresponding timestep. The remaining $T\times h \times w$ video tokens are not fed to any loss. Since all layers of TAPNext output $T\times Q$ track tokens, we apply losses at every TAPNext layer with equal weights and use the last layer's output as the prediction. We found this combined loss significantly improved performance. 


Another key innovation of TAPNext is the parameterization of the coordinate head. Specifically, we propose to parameterize the \textit{coordinate prediction as classification of coordinates} (inspired by \citet{farebrother2024stop}). The motivation of this approach is that tracking assumes that the range of coordinates is bounded by the image size. Specifically, $x$ and $y$ coordinates are discretized into $n$ ($n=256$ in our experiments) classifier bins. Each bin corresponds to a certain discrete coordinate. Although the classifier selects among a finite number of discrete coordinates, we can obtain a continuous coordinate prediction by computing \textit{the expected coordinate w.r.t. classifier probabilities}. We include a detailed description of the coordinate head in the Appendix \ref{app:coordinate_head}.
Notably, the ablations in Section \ref{sec:experiments} suggest that the coordinate classification is one of the most important components of TAPNext. The coordinate head is trained using a combination of softmax cross entropy loss and Huber loss (on the continuous prediction). Note that the coordinate prediction means that our model can represent \textit{multimodal} predictions if it is uncertain, unlike prior methods such as TAPIR and Co-Tracker which perform multiple refinement steps on a single trajectory hypothesis per query, and perform local feature sampling around that single hypothesis. The visibility head performs binary classification and is trained with sigmoid binary cross entropy loss.  Like with TAPIR, we also make use of an uncertainty estimate, although unlike TAPIR, we do not need to add any extra loss: we simply mark points as occluded if more than 50\% of the probability mass lies outside of an 8-pixel radius.

\textbf{BootsTAP.}
Training on pseudo-labled real data has proven beneficial across many frameworks~\cite{sun2024refining,bootstap,cotracker3} as a method of bridging the sim-to-real domain gap, and we expect this to be particularly important for TAPNext due to its lack of inductive biases. We closely follow the method proposed in BootsTAP~\cite{bootstap}, which intuitively aims to make the network equivariant to affine transformations of the video, and invariant to corruption. Specifically, we pre-train on sim-only data, and then fine-tune with a student-teacher setup, where the `student' TAPNext model trains on both sim data and real pseudo-labeled by a `teacher' network.  For the real data, the `teacher' makes a prediction on un-corrupted, full-resolution videos, while the `student' must make a prediction on a view which is affine-transformed and corrupted with JPEG artifacts.  The loss is backpropped only to the student; the teacher is updated via an exponential moving average of the student's weights, which means it provides stable predictions and is unlikely to collapse. We refer to TAPNext finetuned on real data as BootsTAPNext.

\textbf{Advantages of TAPNext.}
Although our model contains a recurrent component, coordinate predictions at different time-steps do not necessarily depend on each other as in many auto-regressive models. Such dependence can lead to accumulation of errors causing trackers to drift \citep{pips}. Many prior works perform iterative refinement that is bidirectional in time potentially using information from future frames to correct past errors. TAPNext demonstrates that it is possible to achieve state-of-the-art tracking quality by causally processing every frame only once. Our model is conceptually simple, in that it contains few hyper-parameters and no tracking-specific inductive biases. This allows it to find its own, unbiased, solution to TAP without being prescribed how motion and appearance should be used. Lastly, because TAPNext is a \emph{linear} recurrent model, it can track all points in a video in parallel over time thanks to parallel inference of SSMs \cite{mamba,s4,s5}.

\begin{table*}[h!]
\centering

\resizebox{\textwidth}{!}{
    \begin{tabular}{@{}lllll|lll|lll|lll@{}}
    \toprule
    \textbf{Method} & \textbf{Latency} & \multicolumn{3}{c}{\textbf{DAVIS First}} & \multicolumn{3}{c}{\textbf{DAVIS Strided}} & \multicolumn{3}{c}{\textbf{Kinetics First}} & \multicolumn{3}{c}{\textbf{Kinetics Strided}}\\ \midrule
     &  & AJ$\uparrow$ & $\delta^{avg}$ $\uparrow$ & OA $\uparrow$ & AJ$\uparrow$ & $\delta^{avg}$ $\uparrow$ & OA $\uparrow$ & AJ$\uparrow$ & $\delta^{avg}$ $\uparrow$ & OA $\uparrow$ & AJ$\uparrow$ & $\delta^{avg}$ $\uparrow$ & OA $\uparrow$ \\ 
     \midrule
     \multicolumn{14}{c}{Models evaluated at $256\times 256$ resolution.} \\
    \midrule
    TAPNet \citep{tapvid} &  & $33.0$ & $48.6$ & $78.8$ & $38.4$ & $53.1$ & $82.3$ & $38.5$ & $54.4$ & $80.6$ & $46.6$ & $60.9$ & $85.0$ \\
    Online TAPIR \citep{vecerik2024robotap} &  & $56.2$ & $70.0$ & $86.5$ & - & - & - & $51.5$ & - & - & - & - & - \\
    Online BootsTAP \citep{bootstap} &  &  $59.7$ & $72.3$ & $86.9$ & $61.2$ & - & - & $55.1$ & $67.5$ & $86.3$ & - & - & - \\
    Track-On \citep{trackon} &  & $\underline{65.0}$ & $\underline{78.0}$ & $90.8$ & - & - & - & $53.9$ & $67.3$ & $87.8$ & - & - & - \\
    \textbf{TAPNext-S (Ours)} & \textbf{frame} & $59.9$ & $74.4$ & $89.8$ & $60.0$ & $76.6$ & $84.3$ & $49.8$ & $65.6$ & $85.9$ & $52.8$ & $69.5$ & $82.3$  \\
    \textbf{TAPNext-B (Ours)} & & $62.4$ & 76.6 & $90.5$ & $65.4$ & $79.7$ & $88.9$ & $53.3$ & $67.9$ & $87.0$ & $56.6$ & $71.4$ & $84.9$ \\
    \textbf{BootsTAPNext-B (Ours)}  & & $\mathbf{65.2}$ & {$\mathbf{78.5}$} & $\underline{91.2}$ & $\underline{68.9}$ & $\mathbf{82.4}$ & $\mathbf{91.6}$ & $\mathbf{57.3}$ & $\mathbf{70.6}$ & $87.4$ & $\mathbf{62.2}$ & $\mathbf{75.1}$ & $89.0$ \\
    \midrule 
    TAPIR \citep{tapir} &  & $58.5$ & $70.0$ & $86.5$ & $61.3$ & $73.6$ & $88.8$ & $49.6$ & $64.2$ & $85.0$ & $57.2$ & $70.1$ & $87.8$ \\
    BootsTAP \citep{bootstap}&  & $61.4$ & $74.0$ & $88.4$ & $66.2$ & $78.5$ & $90.7$ & $54.6$ & $68.4$ & $86.5$ & $\underline{61.4}$ & $\underline{74.2}$ & $\mathbf{89.7}$ \\
    TAPTR \citep{li2024taptr} & \textbf{window} & $63.0$ & $76.1$ & $91.1$ & $66.3$ & $79.2$ & $91.0$ & $49.0$ & $64.4$ & $85.2$ & - & - & -\\
    TAPTRv2 \citep{li2024taptrv2} & & 63.5 & $75.9$ & $\textbf{91.4}$ & $66.4$ & $78.8$ & $\underline{91.3}$ & $49.7$ & $64.2$ & $85.7$ & - & - & -\\ 
    TAPTRv3 \citep{Qu2024taptrv3} & & $63.2$ & $76.7$ & $91.0$ & - & - & - & $54.5$ & $67.5$ & $\underline{88.2}$ & - & - & -\\ 
    \midrule
    OmniMotion \citep{wang2023omnimotion} &  & - & - & - & $51.7$ & $67.5$ & $85.3$ & - & - & - & $55.1$ & $69.2$ & $\underline{89.2} $\\
    Dino-Tracker \citep{dinotracker} & \textbf{video} & -  & -  & - & $62.3$ & $78.2$ & $87.5$ & - & - & - & $59.7$ & $73.3$ & $88.5$ \\
    LocoTrack-B \cite{cho2024local} & & $63.0$ & $75.3$ & $87.2$ & $67.8$ & $79.6$ & $89.9$ & $52.9$ & $66.8$ & $85.3$ & $59.5$ & $73.0$ & $88.5$\\
    \bottomrule
    \multicolumn{14}{c}{Models evaluated at $384\times 512$ resolution.} \\
    \midrule
    PIPs \citep{pips} &  &  $42.2$ & $64.8$ & $77.7$ & $52.4$ & $70.0$ & $83.6$ & - & - & - & $35.3$ & $54.8$ & $77.4$ \\
    CoTracker2 \citep{cotracker} & \textbf{window} & $62.2$ & $75.7$ & $89.3$ & $65.9$ & $79.4$ & $89.9$ & $48.8$ & $64.5$ & $85.8$ & $57.3$ & $70.6$ & $87.5$ \\ 
    CoTracker3 \citep{cotracker3} &  & $64.4$ & $76.9$ & $\underline{91.2}$ & - & - & - & $54.7$ & $67.8$ & $87.4$ & - & - & - \\
    \midrule
    LocoTrack-B \cite{cho2024local} & \textbf{video} & $64.8$ & $77.4$ & $86.2$ & $\mathbf{69.4}$ & $\underline{81.3}$ & $88.6$ & $52.3$ & $66.4$ & $82.1$ & $59.1$ & $72.5$ & $85.7$ \\
     CoTracker3 \citep{cotracker3} & & $63.8$ & $76.3$ & $90.2$ & - & - & - & $\underline{55.8}$ & $\underline{68.5}$ & $\mathbf{88.3}$ & - & - & - \\ 
    \bottomrule
    \end{tabular}
}
\caption{Tracking performance for TAPNext and baseline models. TAPNext achieves a new state-of-the-art point tracking performance on eight of the twelve metrics, while also achieving the lowest possible latency. Methods are organized by their latency. \textbf{Latency: video} - these models require the entire video as input before outputting the point tracks. \textbf{Latency: window} - these models output tracks of length $T$ after consuming a chunk of $T$ frames (typically $T=8$). After filling a buffer of $T$ frames, these models can operate in a per-frame fashion. \textbf{Latency: frame}  - these models have minimal latency by outputting point predictions immediately after consuming each frame. In each column, the best performing values are in \textbf{bold}, the second best are \underline{underlined}. 
}
\label{table:tapvid_results}
   
\end{table*}

\subsection{Training \& Inference Details}
Like most previous works in TAP, we train and evaluate on the TAP-Vid benchmark \cite{tapvid}. This benchmark uses Kubric to generate synthetic training data, and two human-labeled evaluation datasets: DAVIS (30 videos, 24 to 105 frames) and Kinetics ($1150$ videos, $250$ frames each). While most methods use small scale generated Kubric data ($11,000$ videos, $24$ frames each), we found it important to train TAPNext on a significantly larger dataset of $500,000$ videos each consisting of $48$ frames. Also, we generate videos with camera panning and motion blur. Note that the data generation engine is open source (see details on data generation process in Appendix \ref{app:kubric}) and there is no consensus on which data should be used for training. Some methods \citep{tapvid} use vanilla Kubric, while others use motion blur \citep{cotracker,li2024taptr} or panning \citep{tapir}. Our baseline TAPNext model, like most other baselines \citep{li2024taptr,li2024taptrv2,tapir,tapvid,pips}, trains only on synthetic data. The BootsTAPNext model is finetuned on real videos following \citep{bootstap, cotracker3}. In this case, we train on 48-frame clips from the same dataset presented in BootsTAP \citep{bootstap} (obtained courtesy of the authors), consisting of roughly 15M video clips taken from the internet, containing mostly real videos everyday activities and filtered to remove cuts and overlays. Our best performing model is first trained for 300,000 steps on synthetic data (Kubric), followed by self-supervised training on real videos for an additional 1,500 steps. We closely follow the self-supervised training scheme introduced in BootsTAP \cite{bootstap}, however we replace the TAPIR coordinate loss \cite{tapir} with the TAPNext coordinate loss described in Appendix \ref{app:coordinate_head}. 



\textbf{Training Details.} 
We train TAPNext using batches of $256$ videos with length $48$ frames. We sample $256$ point queries for each video. We train two variants of TAPNext: TAPNext-S (56M parameters) and TAPNext-B (194M parameters). We train the model on $256\times256$ resolution. We use a peak learning rate in the cosine decay schedule of $1e^{-3}$ for the S-model and $5e^{-4}$ for the B-model.
All hyperparameters for training the model and model configuration are included in Appendix \ref{app:hypers}.


\textbf{Inference Details.} We evaluate TAPNext models at $256\times 256$ resolution. To perform query-strided evaluation with our causal tracker, we run the video forwards and backwards from every query point (see Appendix \ref{app:details_strided}).
For fair comparison to other methods, we evaluated TAPNext by tracking one query point at a time and included support points (same as CoTracker \citep{cotracker}). See Appendix \ref{app:joint_tracking} for more details about the experimental evaluation of joint tracking with TAPNext. Like CoTracker, we found that performance can be improved by adding local and global support points to the evaluation.

\section{Experimental Results} 
\label{sec:experiments}
We evaluated TAPNext quantitatively on both the DAVIS and Kinetics components of the TAP-Vid benchmark \citep{tapvid}. Because our model avoids tracking-specific inductive biases, we conducted qualitative evaluation to gain intuition about what visual/motion cues the model uses to track points. Finally we conducted several ablations to further understand the most important ingredients for obtaining good performance.

\textbf{Metrics.} In line with prior works in TAP, we report three metrics for evaluation. First, Occlusion Accuracy (OA) is the accuracy of classifying whether the point is visible or not. Second, coordinate accuracy (denoted $\delta^{avg}$) is a fraction of points within a threshold of $1$, $2$, $4$, $8$, $16$ pixels, averaged over all thresholds. Third, Average Jaccard measures occlusion accuracy and coordinate accuracy together.

\begin{table*}[t]
    \centering
    \begin{tabular}{c|l|c|c|c|c}
    Query Points & \multicolumn{1}{c|}{Model} & \multicolumn{2}{c}{Average FPS} & \multicolumn{2}{|c}{Latency (ms)}  \\
    \midrule
    & & H100 & V100  & H100 & V100 \\
    \hline
    & LocoTrack-B ($256\times256$) & \textbf{452} & \textbf{150} & $2210$ & $6600$ \\
256 & CoTracker3 (online) & 102 & 33 & $80$ & $240$ \\
    & \textbf{TAPNext-B (Ours)} & 197 & 70 & $\mathbf{5.05}$ & $\mathbf{14.2}$ \\
    \hline
    & LocoTrack-B ($256\times256$) & \textbf{242} & \textbf{82} & $4130$ & $12000$ \\
512 & CoTracker3 (online) & 69 & 22 & $116$ & $360$ \\
    & \textbf{TAPNext-B (Ours)} & 189 & 55 & $\mathbf{5.26}$ & $\mathbf{18}$ \\
    \hline
    & LocoTrack-B ($256\times256$) & 124 & $\mathbf{43}$ & $8000$ & $22800$ \\
1024 & CoTracker3 (online) & 45 & 14 & $177$ & $576$ \\
    & \textbf{TAPNext-B (Ours)} & $\mathbf{187}$ & $\mathbf{42}$ & $\mathbf{5.33}$ & $\mathbf{23}$ \\
    \hline
    \end{tabular}
\caption{Speed comparison of TAPNext$^2$ to online Cotracker3 running on Nvidia V100 and H100 GPUs. The latency metric is defined as the maximum (worst case) time between passing a frame to the model and receiving predicted points, and it includes the time it takes to fill and process the initial frame buffer. All models are implemented in PyTorch.}
\label{tab:speed}
\end{table*}

\subsection{Performance on TAP-Vid} 

Table~\ref{table:tapvid_results} shows our results on TAP-Vid datasets.  As we are targeting online applications, such as robotics and autonomous driving, 
we group approaches by their latency: in particular, methods that can outputs tracks immediately after ingesting each frame, followed by those require a window of frames before they output the first prediction, and then those that require the entire video to track points \citep{dinotracker, wang2023omnimotion}. We also separately report methods that begin by upsampling the image to higher resolution before tracking.


TAPNext achieves state-of-the-art performance in almost all datasets, with particularly large gaps comparing to other online (1-frame-latency) methods. 
Recall that all videos in the Kinetics dataset have $250$ frames. Since the model was trained on $48$ frame long videos it can generalize to \textit{videos that are at least \textbf{5$\times$} longer videos than the ones seen in training}. TAP evaluation is performed simply by propagating frames and query points through our model, i.e. without SSM state resets, windowed inference, test time optimization, or other heuristics. We note that TAPNext achieves this while being substantially simpler than other methods, avoiding tracking-specific inductive biases in its architecture.

We encourage readers to view the qualitative results we have included in the supplementary file, which includes a side-by-side comparison between BootsTAPNext, BootsTAPIR, and Cotracker3.  In particular, we find that our method can more accurately handle occlusion and fast motion. It is also better at tracking thin/small, and texture-less objects.


\textbf{Tracking Speed.} We evaluate the speed of point tracking of TAPNext, LocoTrack-B, and Cotracker3 in Table \ref{tab:speed}. We report two metrics: speed in \textit{average frames-per-second (FPS)}, which is the total time the tracker spent to process the entire video divided by the total number of frames in that video; and \textit{latency}, which is the worst case delay between receiving a frame and outputting point coordinates/visibility of points in that frame (latency is equal to the time of performing one forward pass for approaches that don't use test-time-optimization). Latency measures how much time the user needs to wait until receiving tracked points from the model when tracking in purely online fashion.
TAPNext operates per-frame, while Cotracker3 requires at least 8 frames to produce an output, and LocoTrack-B requires the entire video (the longest video in DAVIS is 104 frames). Table \ref{tab:speed} reports metrics on a video of $1000$ frames.


\subsection{Model Interpretation}
\label{sec:interp}

\textbf{Visualization.} To gain insight into \emph{how} TAPNext tracks points, we visualize the attention weights in several ViT layers. First we visualize the attention weight between track token and image tokens (Figure \ref{fig:spatial_attn}). Next, we show how points interact in the attention process (Figure \ref{fig:query_attn}). Attention patterns implemented by TAPNext show that it uses motion (motion-cluster head in Figure \ref{fig:spatial_attn} and \ref{fig:query_attn}), coordinate (coordinate head in Figure \ref{fig:spatial_attn}), and appearance-matching (cost-volume-like head in Figure \ref{fig:spatial_attn}) cues to implement tracking. Remarkably, these heuristics were not explicitly encoded in the model's architecture or training process. Instead, they arose organically as emergent properties of the system through end-to-end supervised learning.

\noindent \textbf{Ablations.} Table \ref{tab:ablation1} shows ablations of key components of TAPNext. Note that we perform ablations at a smaller scale compared to the model in Table \ref{table:tapvid_results}. We observe that the most important component of TAPNext is the classification coordinate head that uses both classification and regression loss. Another important aspect is to use a small image patch size of $8\times8$ pixels for the initial linear projection. Finally, Table \ref{tab:ablation2} shows that when the SSM is swapped for temporal attention, the model exhibits poor temporal generalization despite using the RoPE \citep{rope-paper} temporal positional embedding which is known to generalize over time.

\noindent \textbf{Reconstructions.} We can visualize internal representations captured by the model by adding a reconstruction term tied to the $T\times h\times w$ output image tokens, which are not directly connected to any output head in the original model. While we did not find that image reconstruction objective leads to improved tracking performance, we can nevertheless use the reconstructions to gain insight into how TAPNext works. To do this, we \textit{invert} the tracking task - instead of the training on the complete video and incomplete motion information, the input contains complete sequences of ground truth point tracks but only a short prefix of video frames. In this setting, TAPNext is then trained to reconstruct the remaining frames. To do this, we used a trained TAPNext model to track points in natural videos, we used the TAPTube dataset used in BootsTAP \citep{bootstap}, and stored the resulting tracks as pseudolabels. Next, we trained another TAPNext model on TAPTube where the \emph{output image tokens} are used to decode the corresponding image patches in future frames. To do this we simply use a linear read-out head to decode the \emph{masked} image tokens to $8\times8$ image patches and minimize the $L_2$ reconstruction loss to the corresponding ground truth patches. Like masked point tokens, masked image tokens are simply positional embedding plus the fixed learned mask token. We evaluate this model on the DAVIS dataset by visualizing reconstructed future frames conditioned only two initial frames and psuedo- ground truth track tokens. The results are shown in Figure \ref{fig:gen}. This should not be veiewed as a generative model, but rather a \emph{linear probing} experiment which visualizes the information stored in the intermediate tokens used by model. We used the TAPNext-B model variant for this experiment. Figure \ref{fig:gen} in the Appendix show key frames along the video completion. The image tokens corresponding to the first 10 frames are input to the model (top two frames of each Figure), after that only the masked image tokens are input and the images are reconstructed from the image tokens imputed by the model. We observe that the model propagates visual information accurately for regions covered by tracked points. The other visual regions (e.g. the ones that appear after the initial points grid moves away) in video completions are not reconstructed well. This is because the model does not have any previous information about those regions (neither visual nor represented through tracked motion), therefore the model simply fills these regions with average values. This implies that the model maintains an accurate visual representation of the tracked points. In other words, the model learns to propagate visual information to future frames from past frames using the trajectories of tracked points. In particular in Figure \ref{fig:gen} (left column) note the blue road sign behind the car is reconstructed fairly accurately even after it was occluded by the car. This implies that the appearance is stably encoded by the model and effectively propagated forward in time by the SSM layers.\footnotetext{We use \texttt{torch.compile} for the inference with TAPNext as we found it to be crucial for fast inference. We tried our best to also apply compilation for CoTracker3 and LocoTrack but for CoTracker3 compilation makes inference slower and for LocoTracker compilation led to out of memory error.}

\subsection{Behavior for Long Video Lengths}
We observe significant failure in long term point tracking the full length of a video greater than 150 frames. The issue is likely due to the state space model, which is only trained with maximum 48 frames and generalizes poorly to significantly longer video clips. There is large opportunity to further improve the strong tracking numbers even more if future work addresses this limitation in the SSM. We have found a partial mitigation to the temporal degradation to be: clipping the forget gate in the SSM to between 0.0 and 0.1, and broadcasting the query features across the length of the video tokens, so that the query points are not lost in the temporal context.
\label{sec:interp}

\begin{table}[h!]
    \begin{adjustbox}{width=\linewidth,center}
    \begin{tabular}{ccc|c}
        \toprule
        \midrule 
        \multicolumn{3}{c}{\makecell[c]{Type of Ablation: \\ default value $\rightarrow$ ablated value}} & \makecell[c]{Average  \\ Jaccard} \\ 
         \midrule 
        \multicolumn{3}{c}{\makecell[c]{TAPNext-S default (small scale run)}} & $55.0$ \\
        \midrule
         \makecell[c]{Classification \\ coordinate head} & $\rightarrow$ & \makecell[c]{Regression \\ coordinate head} & $44.7$ \\
         \midrule 
        \makecell[c]{$2\times$ state expansion \citep{griffin} \\ in SSM} & $\rightarrow$ & \makecell[c]{No SSM \\ state expansion} & $53.8$ \\
        \midrule 
        \makecell[c]{Losses after each \\ ViT Block} & $\rightarrow$ & \makecell[c]{No intermediate \\ losses} & $50.5$ \\
        \midrule 
        \makecell[c]{Regression $+$ Classification \\ coordinate loss} & $\rightarrow$ & \makecell[c]{Classification \\ coordinate loss} & $52.7$ \\
        \midrule
        \makecell[c]{Regression $+$ Classification \\ coordinate loss} & $\rightarrow$ & \makecell[c]{Regression \\ coordinate loss} & $48.1$ \\
        \midrule
        \makecell[c]{Image patch $8\times 8$} & $\rightarrow$ & \makecell[c]{Image patch $16\times 16$} & $49.7$ \\
        \midrule
        \bottomrule
    \end{tabular}
    \end{adjustbox}
    \caption{Ablating of the main components of TAPNext-S. We train each model for $150,000$ steps and batch size $128$ and on $24$ frames (compared to $300,000$ steps, batch $256$, and $48$ frames for the main TAPNext models in Table \ref{table:tapvid_results}). We evaluate every ablation on DAVIS query-first. All ablations are independent of each other.}
    \label{tab:ablation1}
\end{table}

\begin{table}[h!]
    \vspace{-10pt}
    \begin{adjustbox}{width=\linewidth,center}
    \begin{tabular}{c|ccc}
        \toprule
        \makecell[c]{TAPNext \\ Variants}& \makecell[c]{Average Jaccard \\ 24 Frames} & \makecell[c]{Average Jaccard \\ Full length} \\ 
        \midrule 
        \makecell[c]{Temporal Attention} & $68.4$ & $17.3$ \\
        \makecell[c]{Temporal SSM} & $70.0$ & $55.0$ \\
        \bottomrule
    \end{tabular}
    \end{adjustbox}
    \caption{Temporal attention ablation of TAPNext-S (under the same protocol as Table \ref{tab:ablation1}).
    We change the temporal SSM block to the temporal attention block \citep{gemmateam2024gemmaopenmodelsbased} that uses rotary positional embeddings \citep{rope-paper}. SSM blocks enable much better temporal generalization in TAPNext beyond the 24 frame training sequences.}
    \label{tab:ablation2}
\end{table}

 \section{Conclusion}
We introduced TAPNext, a framework that casts TAP as next token prediction. While being conceptually simple, our method achieves competitive performance. We hope that this will inspire researchers to build on top of TAPNext and further increase the popularity of TAP as an important computer vision task. Although we applied this framework only to point tracking it can be extended to many other computer vision tasks in video.

{
    \small
    \bibliographystyle{ieeenat_fullname}
    \bibliography{main}

\begin{thebibliography}{62}
\providecommand{\natexlab}[1]{#1}
\providecommand{\url}[1]{\texttt{#1}}
\expandafter\ifx\csname urlstyle\endcsname\relax
  \providecommand{\doi}[1]{doi: #1}\else
  \providecommand{\doi}{doi: \begingroup \urlstyle{rm}\Url}\fi

\bibitem[Aydemir et~al.(2025)Aydemir, Cai, Xie, and G\"uney]{trackon}
G\"orkay Aydemir, Xiongyi Cai, Weidi Xie, and Fatma G\"uney.
\newblock {Track-On}: Transformer-based online point tracking with memory.
\newblock In \emph{The Thirteenth International Conference on Learning Representations}, 2025.

\bibitem[Bachmann et~al.(2022)Bachmann, Mizrahi, Atanov, and Zamir]{bachmann2022multimae}
Roman Bachmann, David Mizrahi, Andrei Atanov, and Amir Zamir.
\newblock {MultiMAE}: Multi-modal multi-task masked autoencoders.
\newblock In \emph{European Conference on Computer Vision}, pages 348--367. Springer, 2022.

\bibitem[Baker and Matthews(2004)]{baker2004lucas}
Simon Baker and Iain Matthews.
\newblock Lucas-kanade 20 years on: A unifying framework.
\newblock \emph{International journal of computer vision}, 56:\penalty0 221--255, 2004.

\bibitem[Balasingam et~al.(2024)Balasingam, Chandler, Li, Zhang, and Balakrishnan]{balasingam2024drivetrack}
Arjun Balasingam, Joseph Chandler, Chenning Li, Zhoutong Zhang, and Hari Balakrishnan.
\newblock Drivetrack: A benchmark for long-range point tracking in real-world videos.
\newblock In \emph{Proceedings of the IEEE/CVF Conference on Computer Vision and Pattern Recognition}, pages 22488--22497, 2024.

\bibitem[Bharadhwaj et~al.(2024)Bharadhwaj, Mottaghi, Gupta, and Tulsiani]{bharadhwaj2024track2act}
Homanga Bharadhwaj, Roozbeh Mottaghi, Abhinav Gupta, and Shubham Tulsiani.
\newblock {Track2Act}: Predicting point tracks from internet videos enables diverse zero-shot robot manipulation.
\newblock \emph{arXiv preprint arXiv:2405.01527}, 2024.

\bibitem[Bharadhwaj et~al.(2025)Bharadhwaj, Dwibedi, Gupta, Tulsiani, Doersch, Xiao, Shah, Xia, Sadigh, and Kirmani]{bharadhwaj2024gen2act}
Homanga Bharadhwaj, Debidatta Dwibedi, Abhinav Gupta, Shubham Tulsiani, Carl Doersch, Ted Xiao, Dhruv Shah, Fei Xia, Dorsa Sadigh, and Sean Kirmani.
\newblock Gen2act: Human video generation in novel scenarios enables generalizable robot manipulation.
\newblock \emph{CoRL Workshop on X-Embodiment}, 2025.

\bibitem[Botev et~al.(2024)Botev, De, Smith, Fernando, Muraru, Haroun, Berrada, Pascanu, Sessa, Dadashi, Hussenot, Ferret, Girgin, Bachem, Andreev, Kenealy, Mesnard, Hardin, Bhupatiraju, Pathak, Sifre, Rivière, Kale, Love, Tafti, Joulin, Fiedel, Senter, Chen, Srinivasan, Desjardins, Budden, Doucet, Vikram, Paszke, Gale, Borgeaud, Chen, Brock, Paterson, Brennan, Risdal, Gundluru, Devanathan, Mooney, Chauhan, Culliton, Martins, Bandy, Huntsperger, Cameron, Zucker, Warkentin, Peran, Giang, Ghahramani, Farabet, Kavukcuoglu, Hassabis, Hadsell, Teh, and de~Frietas]{botev2024recurrentgemma}
Aleksandar Botev, Soham De, Samuel~L Smith, Anushan Fernando, George-Cristian Muraru, Ruba Haroun, Leonard Berrada, Razvan Pascanu, Pier~Giuseppe Sessa, Robert Dadashi, Léonard Hussenot, Johan Ferret, Sertan Girgin, Olivier Bachem, Alek Andreev, Kathleen Kenealy, Thomas Mesnard, Cassidy Hardin, Surya Bhupatiraju, Shreya Pathak, Laurent Sifre, Morgane Rivière, Mihir~Sanjay Kale, Juliette Love, Pouya Tafti, Armand Joulin, Noah Fiedel, Evan Senter, Yutian Chen, Srivatsan Srinivasan, Guillaume Desjardins, David Budden, Arnaud Doucet, Sharad Vikram, Adam Paszke, Trevor Gale, Sebastian Borgeaud, Charlie Chen, Andy Brock, Antonia Paterson, Jenny Brennan, Meg Risdal, Raj Gundluru, Nesh Devanathan, Paul Mooney, Nilay Chauhan, Phil Culliton, Luiz~Gustavo Martins, Elisa Bandy, David Huntsperger, Glenn Cameron, Arthur Zucker, Tris Warkentin, Ludovic Peran, Minh Giang, Zoubin Ghahramani, Clément Farabet, Koray Kavukcuoglu, Demis Hassabis, Raia Hadsell, Yee~Whye Teh, and Nando de Frietas.
\newblock {RecurrentGemma}: Moving past transformers for efficient open language models, 2024.

\bibitem[Carion et~al.(2020)Carion, Massa, Synnaeve, Usunier, Kirillov, and Zagoruyko]{carion2020detr}
Nicolas Carion, Francisco Massa, Gabriel Synnaeve, Nicolas Usunier, Alexander Kirillov, and Sergey Zagoruyko.
\newblock End-to-end object detection with transformers, 2020.

\bibitem[Cho et~al.(2024)Cho, Huang, Nam, An, Kim, and Lee]{cho2024local}
Seokju Cho, Jiahui Huang, Jisu Nam, Honggyu An, Seungryong Kim, and Joon-Young Lee.
\newblock Local all-pair correspondence for point tracking.
\newblock \emph{arXiv preprint arXiv:2407.15420}, 2024.

\bibitem[De et~al.(2024)De, Smith, Fernando, Botev, Cristian-Muraru, Gu, Haroun, Berrada, Chen, Srinivasan, Desjardins, Doucet, Budden, Teh, Pascanu, Freitas, and Gulcehre]{griffin}
Soham De, Samuel~L. Smith, Anushan Fernando, Aleksandar Botev, George Cristian-Muraru, Albert Gu, Ruba Haroun, Leonard Berrada, Yutian Chen, Srivatsan Srinivasan, Guillaume Desjardins, Arnaud Doucet, David Budden, Yee~Whye Teh, Razvan Pascanu, Nando~De Freitas, and Caglar Gulcehre.
\newblock Griffin: Mixing gated linear recurrences with local attention for efficient language models, 2024.

\bibitem[Doersch et~al.(2022)Doersch, Gupta, Markeeva, Recasens, Smaira, Aytar, Carreira, Zisserman, and Yang]{tapvid}
Carl Doersch, Ankush Gupta, Larisa Markeeva, Adria Recasens, Lucas Smaira, Yusuf Aytar, Joao Carreira, Andrew Zisserman, and Yi Yang.
\newblock {TAP}-vid: A benchmark for tracking any point in a video.
\newblock \emph{Advances in Neural Information Processing Systems}, 35:\penalty0 13610--13626, 2022.

\bibitem[Doersch et~al.(2023)Doersch, Yang, Vecerik, Gokay, Gupta, Aytar, Carreira, and Zisserman]{tapir}
Carl Doersch, Yi Yang, Mel Vecerik, Dilara Gokay, Ankush Gupta, Yusuf Aytar, Joao Carreira, and Andrew Zisserman.
\newblock {TAPIR}: Tracking any point with per-frame initialization and temporal refinement.
\newblock In \emph{Proceedings of the IEEE/CVF International Conference on Computer Vision}, pages 10061--10072, 2023.

\bibitem[Doersch et~al.(2024)Doersch, Luc, Yang, Gokay, Koppula, Gupta, Heyward, Rocco, Goroshin, Carreira, and Zisserman]{bootstap}
Carl Doersch, Pauline Luc, Yi Yang, Dilara Gokay, Skanda Koppula, Ankush Gupta, Joseph Heyward, Ignacio Rocco, Ross Goroshin, João Carreira, and Andrew Zisserman.
\newblock {BootsTAP}: Bootstrapped training for tracking any point.
\newblock \emph{arXiv}, 2024.

\bibitem[Dosovitskiy et~al.(2015)Dosovitskiy, Fischer, Ilg, Hausser, Hazirbas, Golkov, Van Der~Smagt, Cremers, and Brox]{dosovitskiy2015flownet}
Alexey Dosovitskiy, Philipp Fischer, Eddy Ilg, Philip Hausser, Caner Hazirbas, Vladimir Golkov, Patrick Van Der~Smagt, Daniel Cremers, and Thomas Brox.
\newblock Flownet: Learning optical flow with convolutional networks.
\newblock In \emph{Proceedings of the IEEE international conference on computer vision}, pages 2758--2766, 2015.

\bibitem[Dosovitskiy et~al.(2021)Dosovitskiy, Beyer, Kolesnikov, Weissenborn, Zhai, Unterthiner, Dehghani, Minderer, Heigold, Gelly, Uszkoreit, and Houlsby]{dosovitskiy2021imageworth16x16words}
Alexey Dosovitskiy, Lucas Beyer, Alexander Kolesnikov, Dirk Weissenborn, Xiaohua Zhai, Thomas Unterthiner, Mostafa Dehghani, Matthias Minderer, Georg Heigold, Sylvain Gelly, Jakob Uszkoreit, and Neil Houlsby.
\newblock An image is worth 16x16 words: Transformers for image recognition at scale, 2021.

\bibitem[Farebrother et~al.(2024)Farebrother, Orbay, Vuong, Taiga, Chebotar, Xiao, Irpan, Levine, Castro, Faust, Kumar, and Agarwal]{farebrother2024stop}
Jesse Farebrother, Jordi Orbay, Quan Vuong, Adrien~Ali Taiga, Yevgen Chebotar, Ted Xiao, Alex Irpan, Sergey Levine, Pablo~Samuel Castro, Aleksandra Faust, Aviral Kumar, and Rishabh Agarwal.
\newblock Stop regressing: Training value functions via classification for scalable deep {RL}.
\newblock In \emph{Forty-first International Conference on Machine Learning}, 2024.

\bibitem[Gordon et~al.(2018)Gordon, Farhadi, and Fox]{gordon2018re3}
Daniel Gordon, Ali Farhadi, and Dieter Fox.
\newblock Re3: Real-time recurrent regression networks for visual tracking of generic objects.
\newblock \emph{IEEE Robotics Autom. Lett.}, 3\penalty0 (2):\penalty0 788--795, 2018.

\bibitem[Greff et~al.(2022)Greff, Belletti, Beyer, Doersch, Du, Duckworth, Fleet, Gnanapragasam, Golemo, Herrmann, et~al.]{greff2022kubric}
Klaus Greff, Francois Belletti, Lucas Beyer, Carl Doersch, Yilun Du, Daniel Duckworth, David~J Fleet, Dan Gnanapragasam, Florian Golemo, Charles Herrmann, et~al.
\newblock Kubric: A scalable dataset generator.
\newblock In \emph{Proceedings of the IEEE/CVF conference on computer vision and pattern recognition}, pages 3749--3761, 2022.

\bibitem[Gu and Dao(2024)]{mamba}
Albert Gu and Tri Dao.
\newblock Mamba: Linear-time sequence modeling with selective state spaces, 2024.

\bibitem[Gu et~al.(2022)Gu, Goel, and Ré]{s4}
Albert Gu, Karan Goel, and Christopher Ré.
\newblock Efficiently modeling long sequences with structured state spaces, 2022.

\bibitem[Harley et~al.(2022)Harley, Fang, and Fragkiadaki]{pips}
Adam~W. Harley, Zhaoyuan Fang, and Katerina Fragkiadaki.
\newblock Particle video revisited: {T}racking through occlusions using point trajectories.
\newblock In \emph{ECCV}, 2022.

\bibitem[Hartley and Zisserman(2003)]{hartley2003multiple}
Richard Hartley and Andrew Zisserman.
\newblock \emph{Multiple view geometry in computer vision}.
\newblock Cambridge university press, 2003.

\bibitem[He et~al.(2022)He, Chen, Xie, Li, Doll{\'a}r, and Girshick]{he2022masked}
Kaiming He, Xinlei Chen, Saining Xie, Yanghao Li, Piotr Doll{\'a}r, and Ross Girshick.
\newblock Masked autoencoders are scalable vision learners.
\newblock In \emph{Proceedings of the IEEE/CVF conference on computer vision and pattern recognition}, pages 16000--16009, 2022.

\bibitem[Hochreiter(1997)]{hochreiter1997long}
S Hochreiter.
\newblock Long short-term memory.
\newblock \emph{Neural Computation MIT-Press}, 1997.

\bibitem[Huang and Netravali(1994)]{huang1994motion}
Thomas~S Huang and Arun~N Netravali.
\newblock Motion and structure from feature correspondences: A review.
\newblock \emph{Proceedings of the IEEE}, 82\penalty0 (2):\penalty0 252--268, 1994.

\bibitem[Karaev et~al.(2023)Karaev, Rocco, Graham, Neverova, Vedaldi, and Rupprecht]{cotracker}
Nikita Karaev, Ignacio Rocco, Benjamin Graham, Natalia Neverova, Andrea Vedaldi, and Christian Rupprecht.
\newblock {CoTracker}: It is better to track together.
\newblock 2023.

\bibitem[Karaev et~al.(2024)Karaev, Makarov, Wang, Neverova, Vedaldi, and Rupprecht]{cotracker3}
Nikita Karaev, Iurii Makarov, Jianyuan Wang, Natalia Neverova, Andrea Vedaldi, and Christian Rupprecht.
\newblock {CoTracker3}: Simpler and better point tracking by pseudo-labelling real videos.
\newblock \emph{arXiv preprint arXiv:2410.11831}, 2024.

\bibitem[Karpathy and Fei-Fei(2015)]{karpathy2015deep}
Andrej Karpathy and Li Fei-Fei.
\newblock Deep visual-semantic alignments for generating image descriptions.
\newblock In \emph{Proceedings of the IEEE conference on computer vision and pattern recognition}, pages 3128--3137, 2015.

\bibitem[Kirillov et~al.(2023)Kirillov, Mintun, Ravi, Mao, Rolland, Gustafson, Xiao, Whitehead, Berg, Lo, et~al.]{kirillov2023segment}
Alexander Kirillov, Eric Mintun, Nikhila Ravi, Hanzi Mao, Chloe Rolland, Laura Gustafson, Tete Xiao, Spencer Whitehead, Alexander~C Berg, Wan-Yen Lo, et~al.
\newblock Segment anything.
\newblock In \emph{Proceedings of the IEEE/CVF International Conference on Computer Vision}, pages 4015--4026, 2023.

\bibitem[Koppula et~al.(2024)Koppula, Rocco, Yang, Heyward, Carreira, Zisserman, Brostow, and Doersch]{koppula2024tapvid}
Skanda Koppula, Ignacio Rocco, Yi Yang, Joe Heyward, Jo{\~a}o Carreira, Andrew Zisserman, Gabriel Brostow, and Carl Doersch.
\newblock Tapvid-3d: A benchmark for tracking any point in 3d.
\newblock \emph{NeurIPS Datasets and Benchmarks}, 2024.

\bibitem[Kumar et~al.(2025)Kumar, Padmanabhan, Luo, Rambhatla, and Shrivastava]{kumar2025trajectory}
Pulkit Kumar, Namitha Padmanabhan, Luke Luo, Sai~Saketh Rambhatla, and Abhinav Shrivastava.
\newblock Trajectory-aligned space-time tokens for few-shot action recognition.
\newblock In \emph{European Conference on Computer Vision}, pages 474--493. Springer, 2025.

\bibitem[Kwak et~al.(2015)Kwak, Cho, Laptev, Ponce, and Schmid]{kwak2015unsupervised}
Suha Kwak, Minsu Cho, Ivan Laptev, Jean Ponce, and Cordelia Schmid.
\newblock Unsupervised object discovery and tracking in video collections.
\newblock In \emph{Proceedings of the IEEE international conference on computer vision}, pages 3173--3181, 2015.

\bibitem[Li et~al.(2024{\natexlab{a}})Li, Zhang, Liu, Zeng, Li, Ren, Li, and Zhang]{li2024taptrv2}
Hongyang Li, Hao Zhang, Shilong Liu, Zhaoyang Zeng, Feng Li, Tianhe Ren, Bohan Li, and Lei Zhang.
\newblock Taptrv2: Attention-based position update improves tracking any point, 2024{\natexlab{a}}.

\bibitem[Li et~al.(2024{\natexlab{b}})Li, Zhang, Liu, Zeng, Ren, Li, and Zhang]{li2024taptr}
Hongyang Li, Hao Zhang, Shilong Liu, Zhaoyang Zeng, Tianhe Ren, Feng Li, and Lei Zhang.
\newblock Taptr: Tracking any point with transformers as detection, 2024{\natexlab{b}}.

\bibitem[Li et~al.(2024{\natexlab{c}})Li, Li, Wang, He, Wang, Wang, and Qiao]{li2024videomamba}
Kunchang Li, Xinhao Li, Yi Wang, Yinan He, Yali Wang, Limin Wang, and Yu Qiao.
\newblock Videomamba: State space model for efficient video understanding.
\newblock \emph{arXiv preprint arXiv:2403.06977}, 2024{\natexlab{c}}.

\bibitem[Li et~al.(2024{\natexlab{d}})Li, Zhang, Chen, Matusik, Liu, Rus, and Sitzmann]{li2024unifying}
Sizhe~Lester Li, Annan Zhang, Boyuan Chen, Hanna Matusik, Chao Liu, Daniela Rus, and Vincent Sitzmann.
\newblock Unifying 3d representation and control of diverse robots with a single camera.
\newblock \emph{arXiv preprint arXiv:2407.08722}, 2024{\natexlab{d}}.

\bibitem[Li et~al.(2024{\natexlab{e}})Li, Wang, Zhang, Wang, Yuan, Xie, Zou, and Shan]{li2024image}
Yaowei Li, Xintao Wang, Zhaoyang Zhang, Zhouxia Wang, Ziyang Yuan, Liangbin Xie, Yuexian Zou, and Ying Shan.
\newblock Image conductor: Precision control for interactive video synthesis.
\newblock \emph{arXiv preprint arXiv:2406.15339}, 2024{\natexlab{e}}.

\bibitem[Polajnar et~al.(2024)Polajnar, Kvinikadze, Harley, and Malenovsk{\`y}]{polajnar2024wing}
Jernej Polajnar, Elizaveta Kvinikadze, Adam~W Harley, and Igor Malenovsk{\`y}.
\newblock Wing buzzing as a mechanism for generating vibrational signals in psyllids.
\newblock \emph{Insect Science}, 2024.

\bibitem[Pătrăucean et~al.(2023)Pătrăucean, Smaira, Gupta, Continente, Markeeva, Banarse, Koppula, Heyward, Malinowski, Yang, Doersch, Matejovicova, Sulsky, Miech, Frechette, Klimczak, Koster, Zhang, Winkler, Aytar, Osindero, Damen, Zisserman, and Carreira]{patraucean2023perception}
Viorica Pătrăucean, Lucas Smaira, Ankush Gupta, Adrià~Recasens Continente, Larisa Markeeva, Dylan Banarse, Skanda Koppula, Joseph Heyward, Mateusz Malinowski, Yi Yang, Carl Doersch, Tatiana Matejovicova, Yury Sulsky, Antoine Miech, Alex Frechette, Hanna Klimczak, Raphael Koster, Junlin Zhang, Stephanie Winkler, Yusuf Aytar, Simon Osindero, Dima Damen, Andrew Zisserman, and João Carreira.
\newblock Perception test: A diagnostic benchmark for multimodal video models.
\newblock In \emph{Advances in Neural Information Processing Systems}, 2023.

\bibitem[Pătrăucean et~al.(2024)Pătrăucean, He, Heyward, Zhang, Sajjadi, Muraru, Zholus, Karami, Goroshin, Chen, Osindero, Carreira, and Pascanu]{patraucean2024trecvitrecurrentvideotransformer}
Viorica Pătrăucean, Xu~Owen He, Joseph Heyward, Chuhan Zhang, Mehdi S.~M. Sajjadi, George-Cristian Muraru, Artem Zholus, Mahdi Karami, Ross Goroshin, Yutian Chen, Simon Osindero, João Carreira, and Razvan Pascanu.
\newblock {TRecViT: A Recurrent Video Transformer}, 2024.

\bibitem[Qu et~al.(2024)Qu, Li, Liu, Zeng, Ren, and Zhang]{Qu2024taptrv3}
Jinyuan Qu, Hongyang Li, Shilong Liu, Zhaoyang Zeng, Tianhe Ren, and Lei Zhang.
\newblock {TAPTRv3: Spatial and Temporal Context Foster Robust Tracking of Any Point in Long Video}.
\newblock \emph{arXiv preprint}, 2024.

\bibitem[Rumelhart et~al.(1986)Rumelhart, Hinton, and Williams]{rumelhart1986learning}
David~E Rumelhart, Geoffrey~E Hinton, and Ronald~J Williams.
\newblock Learning internal representations by error propagation, parallel distributed processing, explorations in the microstructure of cognition, ed. de rumelhart and j. mcclelland. vol. 1. 1986.
\newblock \emph{Biometrika}, 71\penalty0 (599-607):\penalty0 6, 1986.

\bibitem[Schmidt et~al.(2024)Schmidt, Mohareri, DiMaio, and Salcudean]{schmidt2024surgical}
Adam Schmidt, Omid Mohareri, Simon DiMaio, and Septimiu~E Salcudean.
\newblock Surgical tattoos in infrared: A dataset for quantifying tissue tracking and mapping.
\newblock \emph{IEEE Transactions on Medical Imaging}, 2024.

\bibitem[Smith et~al.(2023)Smith, Warrington, and Linderman]{s5}
Jimmy T.~H. Smith, Andrew Warrington, and Scott~W. Linderman.
\newblock Simplified state space layers for sequence modeling, 2023.

\bibitem[Su et~al.(2021)Su, Lu, Pan, Wen, and Liu]{rope-paper}
Jianlin Su, Yu Lu, Shengfeng Pan, Bo Wen, and Yunfeng Liu.
\newblock Roformer: Enhanced transformer with rotary position embedding.
\newblock \emph{arXiv preprint arXiv:2104.09864}, 2021.

\bibitem[Sun et~al.(2024)Sun, Harley, and Guibas]{sun2024refining}
Xinglong Sun, Adam~W Harley, and Leonidas~J Guibas.
\newblock Refining pre-trained motion models.
\newblock In \emph{2024 IEEE International Conference on Robotics and Automation (ICRA)}, 2024.

\bibitem[Szeliski(2022)]{szeliski2022computer}
Richard Szeliski.
\newblock \emph{Computer vision: algorithms and applications}.
\newblock Springer Nature, 2022.

\bibitem[Team et~al.(2024)Team, Mesnard, Hardin, Dadashi, Bhupatiraju, Pathak, Sifre, Rivière, Kale, Love, Tafti, Hussenot, Sessa, Chowdhery, Roberts, Barua, Botev, Castro-Ros, Slone, Héliou, Tacchetti, Bulanova, Paterson, Tsai, Shahriari, Lan, Choquette-Choo, Crepy, Cer, Ippolito, Reid, Buchatskaya, Ni, Noland, Yan, Tucker, Muraru, Rozhdestvenskiy, Michalewski, Tenney, Grishchenko, Austin, Keeling, Labanowski, Lespiau, Stanway, Brennan, Chen, Ferret, Chiu, Mao-Jones, Lee, Yu, Millican, Sjoesund, Lee, Dixon, Reid, Mikuła, Wirth, Sharman, Chinaev, Thain, Bachem, Chang, Wahltinez, Bailey, Michel, Yotov, Chaabouni, Comanescu, Jana, Anil, McIlroy, Liu, Mullins, Smith, Borgeaud, Girgin, Douglas, Pandya, Shakeri, De, Klimenko, Hennigan, Feinberg, Stokowiec, hui Chen, Ahmed, Gong, Warkentin, Peran, Giang, Farabet, Vinyals, Dean, Kavukcuoglu, Hassabis, Ghahramani, Eck, Barral, Pereira, Collins, Joulin, Fiedel, Senter, Andreev, and Kenealy]{gemmateam2024gemmaopenmodelsbased}
Gemma Team, Thomas Mesnard, Cassidy Hardin, Robert Dadashi, Surya Bhupatiraju, Shreya Pathak, Laurent Sifre, Morgane Rivière, Mihir~Sanjay Kale, Juliette Love, Pouya Tafti, Léonard Hussenot, Pier~Giuseppe Sessa, Aakanksha Chowdhery, Adam Roberts, Aditya Barua, Alex Botev, Alex Castro-Ros, Ambrose Slone, Amélie Héliou, Andrea Tacchetti, Anna Bulanova, Antonia Paterson, Beth Tsai, Bobak Shahriari, Charline~Le Lan, Christopher~A. Choquette-Choo, Clément Crepy, Daniel Cer, Daphne Ippolito, David Reid, Elena Buchatskaya, Eric Ni, Eric Noland, Geng Yan, George Tucker, George-Christian Muraru, Grigory Rozhdestvenskiy, Henryk Michalewski, Ian Tenney, Ivan Grishchenko, Jacob Austin, James Keeling, Jane Labanowski, Jean-Baptiste Lespiau, Jeff Stanway, Jenny Brennan, Jeremy Chen, Johan Ferret, Justin Chiu, Justin Mao-Jones, Katherine Lee, Kathy Yu, Katie Millican, Lars~Lowe Sjoesund, Lisa Lee, Lucas Dixon, Machel Reid, Maciej Mikuła, Mateo Wirth, Michael Sharman, Nikolai Chinaev, Nithum Thain, Olivier Bachem,
  Oscar Chang, Oscar Wahltinez, Paige Bailey, Paul Michel, Petko Yotov, Rahma Chaabouni, Ramona Comanescu, Reena Jana, Rohan Anil, Ross McIlroy, Ruibo Liu, Ryan Mullins, Samuel~L Smith, Sebastian Borgeaud, Sertan Girgin, Sholto Douglas, Shree Pandya, Siamak Shakeri, Soham De, Ted Klimenko, Tom Hennigan, Vlad Feinberg, Wojciech Stokowiec, Yu hui Chen, Zafarali Ahmed, Zhitao Gong, Tris Warkentin, Ludovic Peran, Minh Giang, Clément Farabet, Oriol Vinyals, Jeff Dean, Koray Kavukcuoglu, Demis Hassabis, Zoubin Ghahramani, Douglas Eck, Joelle Barral, Fernando Pereira, Eli Collins, Armand Joulin, Noah Fiedel, Evan Senter, Alek Andreev, and Kathleen Kenealy.
\newblock Gemma: Open models based on gemini research and technology, 2024.

\bibitem[Teed and Deng(2020)]{teed2020raft}
Zachary Teed and Jia Deng.
\newblock Raft: Recurrent all-pairs field transforms for optical flow.
\newblock In \emph{Computer Vision--ECCV 2020: 16th European Conference, Glasgow, UK, August 23--28, 2020, Proceedings, Part II 16}, pages 402--419. Springer, 2020.

\bibitem[Tong et~al.(2022)Tong, Song, Wang, and Wang]{tong2022videomae}
Zhan Tong, Yibing Song, Jue Wang, and Limin Wang.
\newblock Videomae: Masked autoencoders are data-efficient learners for self-supervised video pre-training.
\newblock \emph{Advances in neural information processing systems}, 35:\penalty0 10078--10093, 2022.

\bibitem[Tumanyan et~al.(2024)Tumanyan, Singer, Bagon, and Dekel]{dinotracker}
Narek Tumanyan, Assaf Singer, Shai Bagon, and Tali Dekel.
\newblock Dino-tracker: Taming dino for self-supervised point tracking in a single video.
\newblock \emph{arXiv preprint arXiv:2403.14548}, 2024.

\bibitem[Van Den~Oord et~al.(2016)Van Den~Oord, Dieleman, Zen, Simonyan, Vinyals, Graves, Kalchbrenner, Senior, Kavukcuoglu, et~al.]{van2016wavenet}
Aaron Van Den~Oord, Sander Dieleman, Heiga Zen, Karen Simonyan, Oriol Vinyals, Alex Graves, Nal Kalchbrenner, Andrew Senior, Koray Kavukcuoglu, et~al.
\newblock Wavenet: A generative model for raw audio.
\newblock \emph{arXiv preprint arXiv:1609.03499}, 12, 2016.

\bibitem[Vecerik et~al.(2024)Vecerik, Doersch, Yang, Davchev, Aytar, Zhou, Hadsell, Agapito, and Scholz]{vecerik2024robotap}
Mel Vecerik, Carl Doersch, Yi Yang, Todor Davchev, Yusuf Aytar, Guangyao Zhou, Raia Hadsell, Lourdes Agapito, and Jon Scholz.
\newblock Robotap: Tracking arbitrary points for few-shot visual imitation.
\newblock In \emph{2024 IEEE International Conference on Robotics and Automation (ICRA)}, pages 5397--5403. IEEE, 2024.

\bibitem[Wang et~al.(2024)Wang, Karaev, Rupprecht, and Novotny]{wang2023visual}
Jianyuan Wang, Nikita Karaev, Christian Rupprecht, and David Novotny.
\newblock Visual geometry grounded deep structure from motion.
\newblock \emph{CVPR}, 2024.

\bibitem[Wang et~al.(2023)Wang, Chang, Cai, Li, Hariharan, Holynski, and Snavely]{wang2023omnimotion}
Qianqian Wang, Yen-Yu Chang, Ruojin Cai, Zhengqi Li, Bharath Hariharan, Aleksander Holynski, and Noah Snavely.
\newblock Tracking everything everywhere all at once.
\newblock In \emph{International Conference on Computer Vision}, 2023.

\bibitem[Wen et~al.(2023)Wen, Lin, So, Chen, Dou, Gao, and Abbeel]{abbeel}
Chuan Wen, Xingyu Lin, John So, Kai Chen, Qi Dou, Yang Gao, and Pieter Abbeel.
\newblock Any-point trajectory modeling for policy learning.
\newblock \emph{arXiv preprint arXiv:2401.00025}, 2023.

\bibitem[Wu et~al.(2024)Wu, Li, Gu, Zhao, He, Zhang, Shou, Li, Gao, and Zhang]{wu2025draganything}
Weijia Wu, Zhuang Li, Yuchao Gu, Rui Zhao, Yefei He, David~Junhao Zhang, Mike~Zheng Shou, Yan Li, Tingting Gao, and Di Zhang.
\newblock Draganything: Motion control for anything using entity representation.
\newblock In \emph{European Conference on Computer Vision}, pages 331--348. Springer, 2024.

\bibitem[Xiao et~al.(2024)Xiao, Wang, Zhang, Xue, Peng, Shen, and Zhou]{xiao2024spatialtracker}
Yuxi Xiao, Qianqian Wang, Shangzhan Zhang, Nan Xue, Sida Peng, Yujun Shen, and Xiaowei Zhou.
\newblock Spatialtracker: Tracking any 2d pixels in 3d space.
\newblock In \emph{Proceedings of the IEEE/CVF Conference on Computer Vision and Pattern Recognition}, pages 20406--20417, 2024.

\bibitem[Yu et~al.(2023)Yu, Blackburn-Matzen, Nguyen, Wang, Habib~Kazi, and Bousseau]{yu2023videodoodles}
Emilie Yu, Kevin Blackburn-Matzen, Cuong Nguyen, Oliver Wang, Rubaiat Habib~Kazi, and Adrien Bousseau.
\newblock {VideoDoodles}: Hand-drawn animations on videos with scene-aware canvases.
\newblock \emph{ACM TOG}, 42\penalty0 (4):\penalty0 1--12, 2023.

\bibitem[Yuan et~al.(2024)Yuan, Wen, Zhang, and Gao]{yuan2024general}
Chengbo Yuan, Chuan Wen, Tong Zhang, and Yang Gao.
\newblock General flow as foundation affordance for scalable robot learning.
\newblock \emph{arXiv preprint arXiv:2401.11439}, 2024.

\bibitem[Zheng et~al.(2023)Zheng, Harley, Shen, Wetzstein, and Guibas]{zheng2023pointodyssey}
Yang Zheng, Adam~W Harley, Bokui Shen, Gordon Wetzstein, and Leonidas~J Guibas.
\newblock Pointodyssey: A large-scale synthetic dataset for long-term point tracking.
\newblock In \emph{Proceedings of the IEEE/CVF International Conference on Computer Vision}, pages 19855--19865, 2023.

\bibitem[Zhou et~al.(2024)Zhou, Wang, Nie, Lin, Yu, Yu, and Wang]{zhou2024trackgo}
Haitao Zhou, Chuang Wang, Rui Nie, Jinxiao Lin, Dongdong Yu, Qian Yu, and Changhu Wang.
\newblock Trackgo: A flexible and efficient method for controllable video generation.
\newblock \emph{arXiv preprint arXiv:2408.11475}, 2024.

\end{thebibliography}
}

\newpage
{\vskip .375in}
\twocolumn[
\begin{center}
  {{\Large \bf Supplementary material - TAPNext: Tracking Any Point as Next Token Prediction \par}}
  {\vspace*{24pt}}{
    \large
    \lineskip .5em
    \par
  }
  \vskip .5em
  \vspace*{12pt}
\end{center}
]
\setcounter{page}{1}
\setcounter{section}{0}
\renewcommand{\thesection}{\Alph{section}}

\newcommand\Xspace{0.316}
\begin{figure*}
\noindent \begin{minipage}[c]{\textwidth}
    \centering
    \begin{minipage}{\Xspace\textwidth}
    \begin{figure}[H]
        \includegraphics[width=\linewidth]{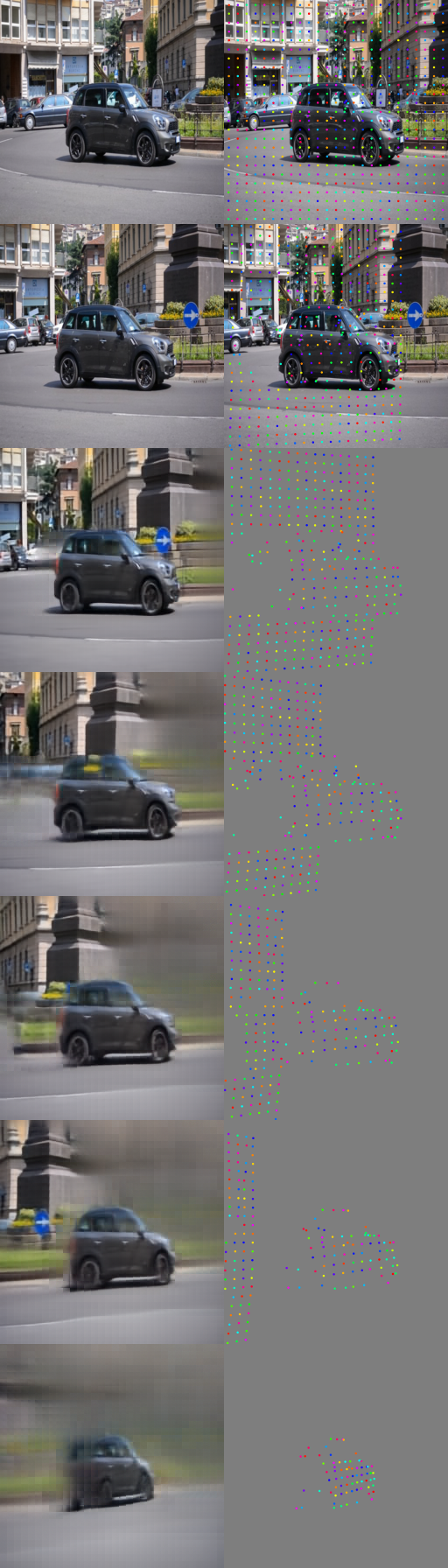}
    \end{figure}
    \end{minipage}
    \begin{minipage}{\Xspace\textwidth}
    \begin{figure}[H]
        \includegraphics[width=\linewidth]{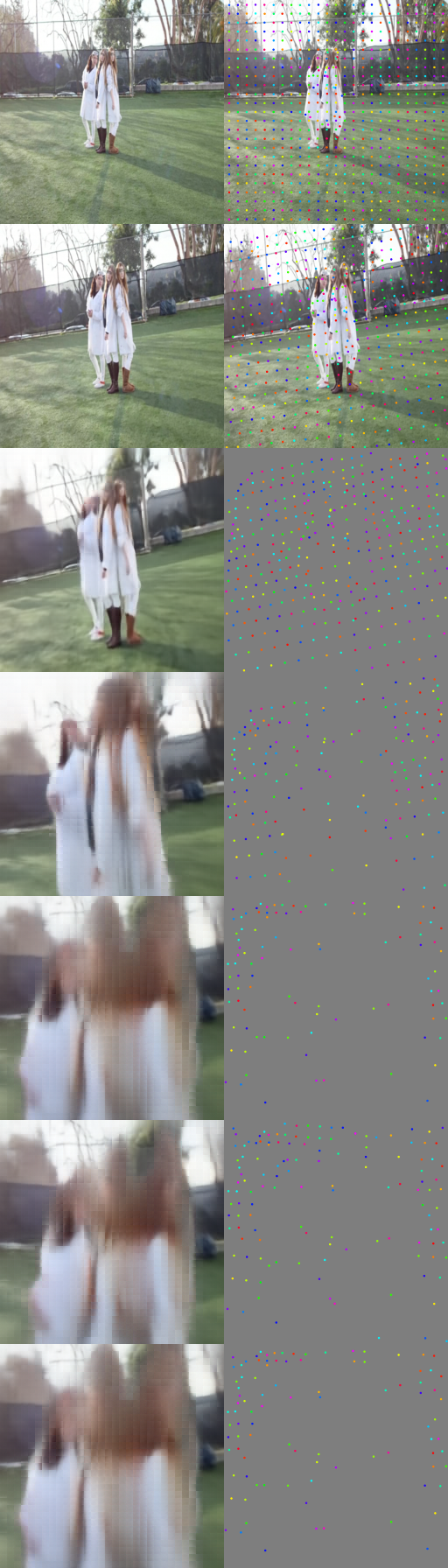}
    \end{figure}
    \end{minipage}
    \begin{minipage}{\Xspace\textwidth}
    \begin{figure}[H]
        \includegraphics[width=\linewidth]{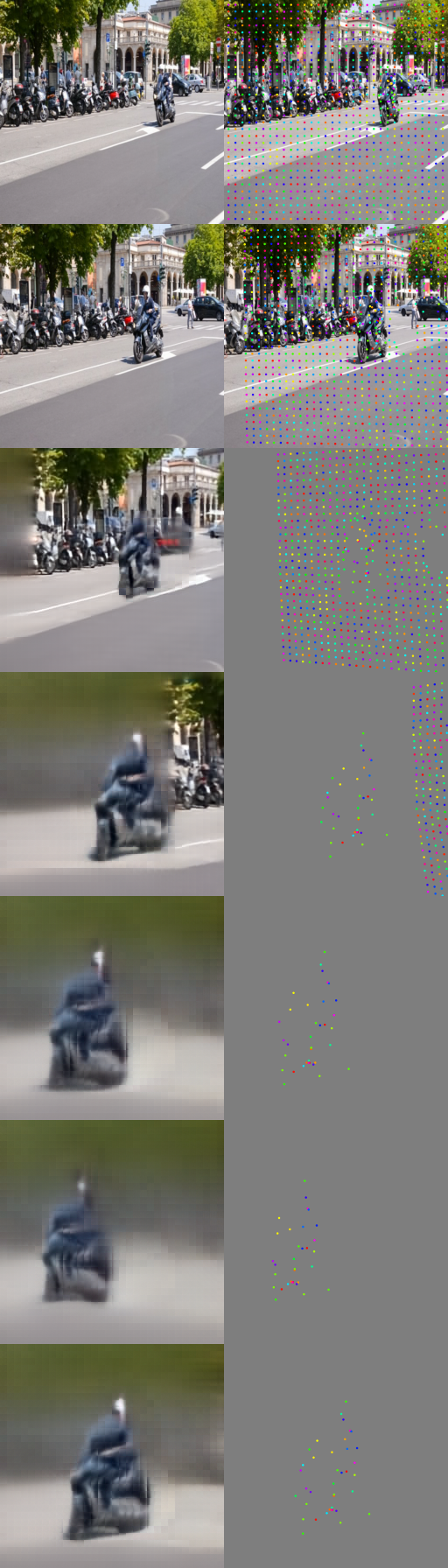}
    \end{figure}
    \end{minipage}
  
\end{minipage}
\vspace{10pt}
\caption{Video Completion by TAPNext variant. \textbf{Left:} Outputs of patch-level linear pixel heads. \textbf{Right:} Inputs to the model (Visible or masked image and points).}
\label{fig:gen}
\end{figure*}
\vspace{20pt}



\section{Frequently Asked Questions}
\label{app:faq}
\begin{itemize}
    \item Is there a reason for categorization of frame/window/video latency? Can't we just run any model e.g. with frame latency (at timestep $t$ we feed frames $1, 2, \cdots, t$ and obtain track prediction right after the frame $t$ was ``fed'')
    \begin{itemize}
        \item This indeed may even turn offline trackers into online ones. However, we would argue that this effectively won't solve the latency problem. The reason is that when such a tracker receives the new frame, it needs to reprocess either all previous frames (for offline trackers) or a window of several recent frames (for window-based trackers). In either case, the time before the prediction is significant effectively disallowing real-time tracking at a high frame rate (see Table \ref{tab:speed} for quantification). 
    \end{itemize}
    \item Why does TAPNext require so much more data and time to train?
    \begin{itemize}
        \item TAPNext is a much more generic model, performing only attention and SSM scanning without any custom components. The generality of TAPNext comes at the cost of much higher compute needed for training (i.e. a large batch and longer optimization). 
    \end{itemize}
    \item TAPNext uses more synthetic and real data to train than previous methods. What is the performance of previous methods given such big datasets remains unclear.
    \begin{itemize}
        \item Currently there is no standartization of which dataset to use, among previous methods. For exampe, TAPIR uses $100000$ videos with camera panning enabled but no motion blur. LocoTrack uses $11000$ synthetic videos for training with panning but no motion blur. TAPTR uses also uses $11000$ training videos which inlude motion blur but don't include camera panning. All aforementioned methods use $24$ frames videos while CoTracker3 trains on synthetic videos of $64$ frames and its training dataset includes only $6000$ training videos at the resolution of $512\times 512$ (compared to $256\times 256$ for previous methods). This way all previous methods use training datasets of different size (which is varied by two orders of magnitude between CoTracker3 and TAPIR), length, resolution and visual properties (motion blur, camera panning).
    \end{itemize}
    \item How does TAPNext learn to recover from occlusions? 
    \begin{itemize}
        \item For each point query TAPNext has the corresponding sequence of SSM recurrent hidden states (since SSM is a form of RNN) that contain the information about the tracked point. Even when the visaul occlusion happens, this information is still being processed by recurrent SSM and the corresponding output predicts the coordinate and occlusion flag. 
    \end{itemize}
    \item Why claiming that the method generalizes to 5x longer videos? This seems to be a weaker statement than prior methods can do given that they track potentially infinite videos.
    \begin{itemize}
    \item TAPNext marks the new stage of point tracking methods with no (tracking) inductive biases that are entirely data driven. On one hand this gives scalability and SOTA tracking quality that comes from the scale of the model and data. On the other hand, due to that it is harder to satisfy guarantees like robustness to long videos. We hope that future research will address this problem of long term tracking of end-to-end trained models. 
    \end{itemize}
    \item Why not putting the TRecViT into the related work? 
    \begin{itemize}
    \item While TRecViT is a strong visual backbone, TAPNext could use other video backbone. Our selection of TRecViT was dictated by its efficiency - not only it requires significantly less time and memory to train (than e.g. a purely attention counterpart), it also processes videos online and needs only one previous recurrent state to perform inference. Due to this reason and since TAPNext's idea is connected to how TRecViT processes tokens, we included the description of the latter.
    \end{itemize}
\end{itemize}

\section{Attention visualization}
\label{app:attn_vis}
For convenience, larger versions of Figures \ref{fig:spatial_attn} and \ref{fig:query_attn}
are reproduced in the appendix as Figures \ref{fig:app:spatial_attn} and \ref{fig:app:query_attn}. We also show the raw spatial attention maps of the attention heads from the same layer in Figure \ref{fig:attn_raw}. Full videos of attention visualization can be found in the supplementary files. Notably in the point-to-point attention video there is strong connection between clusters on different objects, but as the video progresses and the objects (colored in red and blue) move independently these connections quickly disappear. This hints at the model's emergent motion segmentation ability. 

\begin{figure*}
  \begin{center}
    \resizebox{\textwidth}{!}{
        \input{figures/attention_vis/attention}
    }
  \end{center}
  \vspace{-12pt}
  \caption{Three attention patterns learned by TAPNext. We visualize attention maps where the attention queries are the point track tokens and the keys are image tokens, which correspond to $8\times8$ patches. Each row is a certain (layer, head) pair. We observe patterns: \textbf{(top)} Cost-volume-like attention head; \textbf{(middle)} Coordinate-based readout head; \textbf{(bottom)} motion-cluster-based readout head. Note that these are just intermediate heads in the backbone.}
  \label{fig:app:spatial_attn}
\end{figure*}


\begin{figure*}
  \begin{center}
    \resizebox{\textwidth}{!}{
        \input{figures/attention_vis/attnention2}
    }
  \end{center}
  \caption{Point-to-point attention map visualizations. Tracked points are nodes and (scaled) attention weights are edges, the thicker the edge the higher the weight between points. Two frames from a video are used to visualize two attention layers. 
  Note that in all images we see strong attention between points on objects that are moving together.
  }
  \label{fig:app:query_attn}
\end{figure*}

\begin{figure*}
    \centering
    \resizebox{0.8\textwidth}{!}{
         \includegraphics{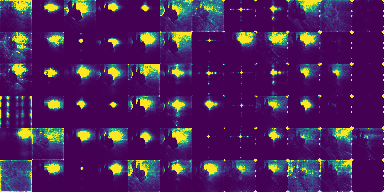}
    }
    \caption{\textbf{X-axis:} layers; \textbf{Y-axis:} attention heads. We visualize attention maps for the TAPNext-S model which has 12 layers and 6 attention heads. We visualize the same video as in Figure \ref{fig:app:spatial_attn}, specifically the timestep corresponding to the leftmost column. Also, we use the same qeury point as in Figure \ref{fig:app:spatial_attn}. The attention maps visualize the attention where query is the track tokens and key is image patch, we visualize it for every head in every layer. The patterns found in Figure \ref{fig:app:spatial_attn} moslty are repeated accross the model.}
    \label{fig:attn_raw}
\end{figure*}

\section{Joint-Tracking and Support Points}
\label{app:joint_tracking}

\begin{table}[h!]
    \centering
    \resizebox{\linewidth}{!}{
    \begin{tabular}{c|c|c|c}
    & AJ & PTS ($\delta^{avg}$) & OA \\
    \hline
    \makecell[c]{Individual Points + 4x4 global \\ + 9x9 local grid} & 62.2 & 76.1 & 90.9  \\
    \midrule
    Joint Points & 62.4 & 76.6 & 90.5 
    \end{tabular}
    }
    \caption{Comparison of joint query point tracking v.s. individual queries and support points on DAVIS first evaluation of TAPNext-B}
    \label{tab:support_table}
\end{table}

\begin{figure*}[b]
    \centering
    \begin{subfigure}{0.5\linewidth}
        \centering
        \includegraphics[width=1.1\linewidth]{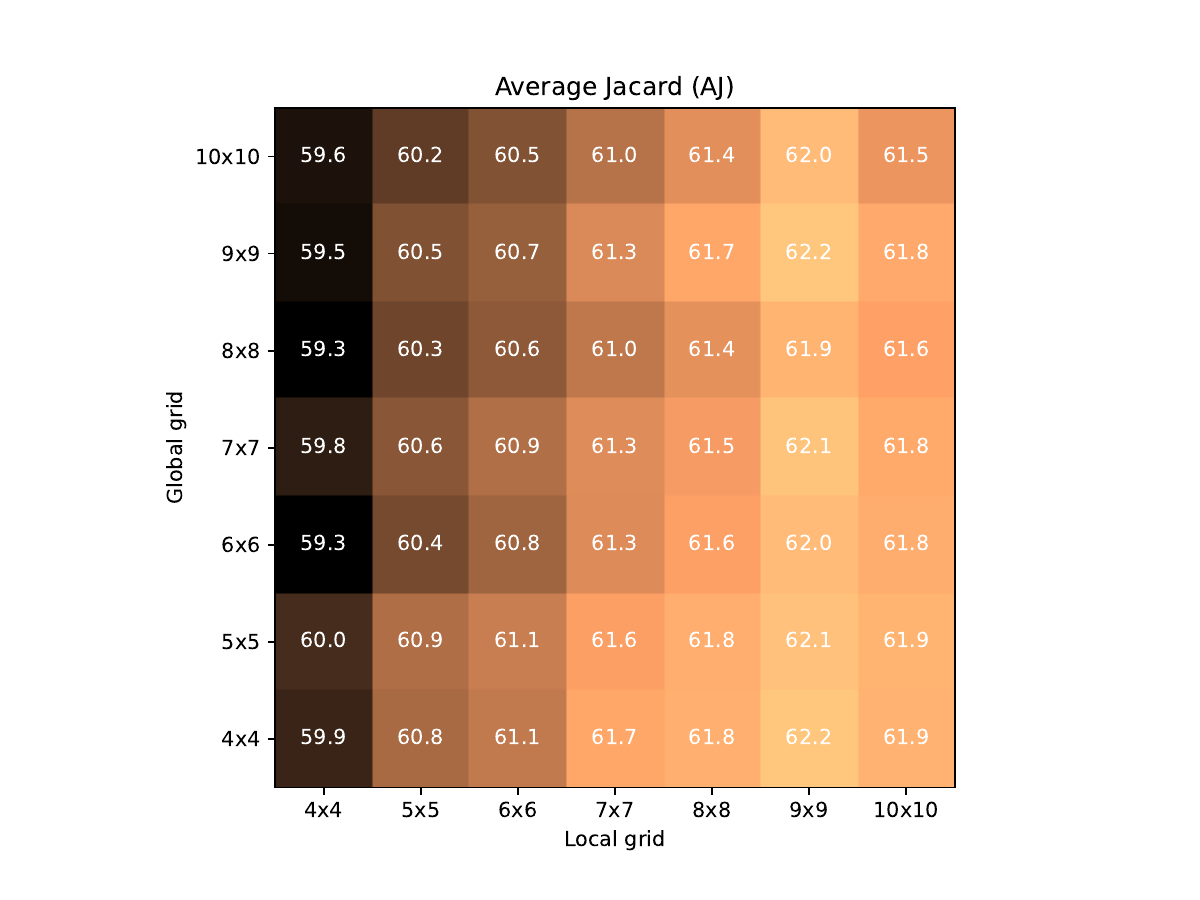}
    \end{subfigure}
    \begin{subfigure}{0.5\linewidth}
        \centering
        \includegraphics[width=1.1\linewidth]{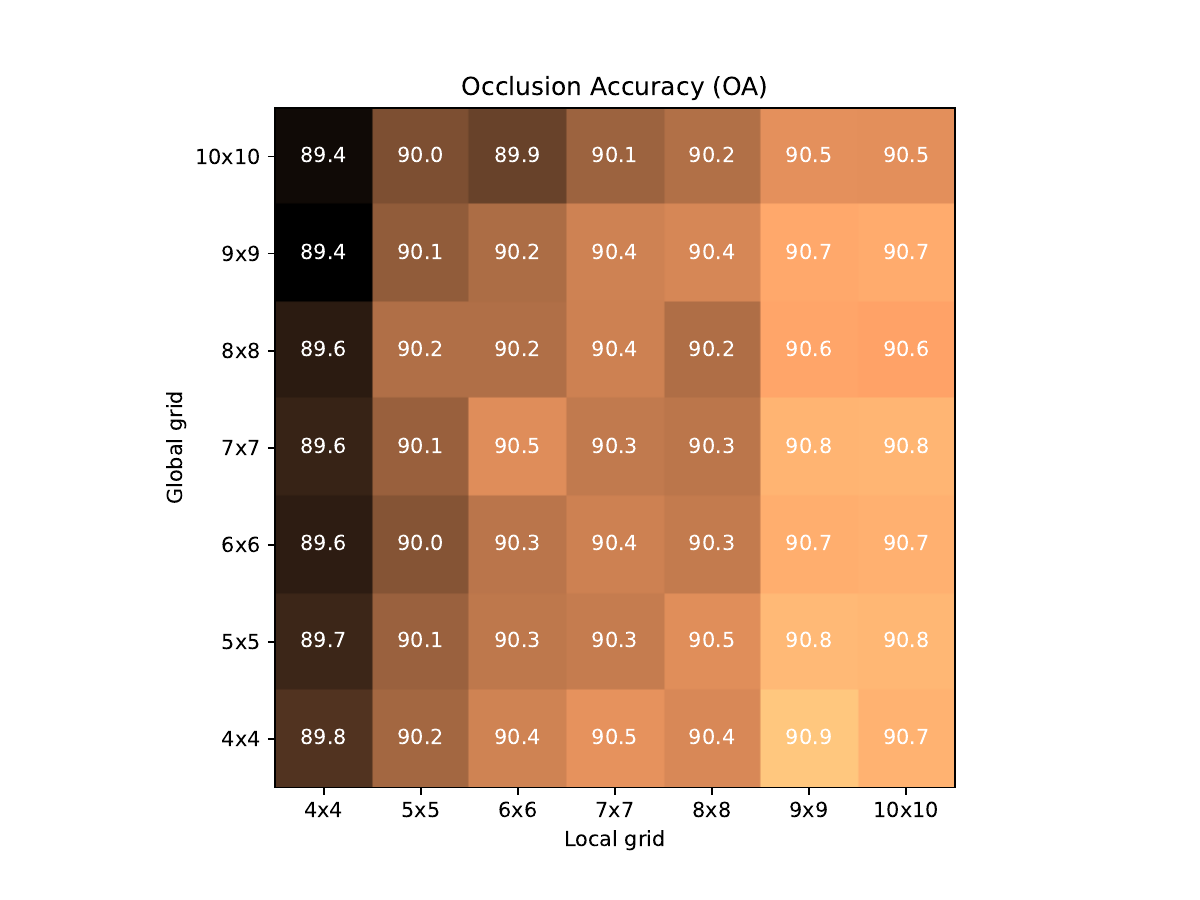}
    \end{subfigure}
    \begin{subfigure}{0.5\linewidth}
        \centering
        \includegraphics[width=1.1\linewidth]{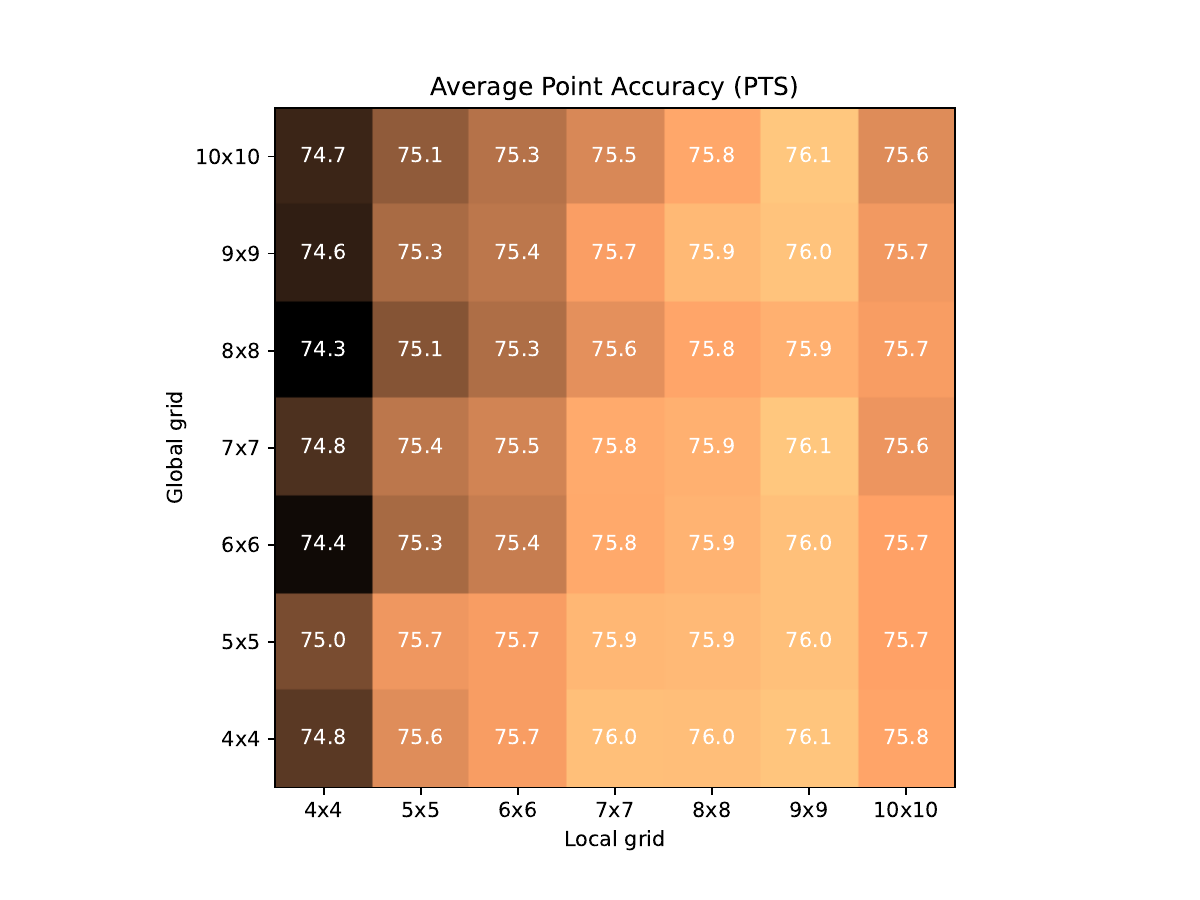}
    \end{subfigure}
\caption{One point at a time tracking performance with various support points grid configurations.}
\label{fig:support_point_sweep}
\end{figure*}

\noindent 
Because TAPNext processes query points jointly there is a concern that semantic correlation between query points can give it an unfair advantage compared to methods that process queries individually. Therefore we use the methodology from Cotracker \citep{cotracker, cotracker3} to evaluate tracking of one point at a time with additional query points sampled from local and global regular grids. Similar to Cotracker we found that TAPNext benefits from mainly from a local support grid of points (Figure \ref{fig:support_point_sweep}). We found no major difference in performance between the one point evaluation (with the best support point scheme) compared to evaluating on all query points jointly, see Table \ref{tab:support_table}.  

\section{Coordinate Prediction via Classification}
\label{app:coordinate_head}

The coordinate prediction head of TAPNext predicts $(x, y) \in [0, H]\times [0, W]$ coordinates.
The $x$ and $y$ coordinates are discretized into $n=256$ values, corresponding to pixel locations, and the head outputs the discrete distribution ($p_i$ where $0 < i < n$, $p \in \mathbb{R}^n$) for both coordinates independently. In particular, each point token is passed to the MLP where the last layer is softmax. To obtain the final, sub-pixel, coordinate we use the truncated soft $argmax$ operation (the same as TAPNet \cite{tapvid}). 
First we truncate the probability distribution around argmax: we set all probability bins that are more than $\Delta$ steps away from argmax to zero ($\Delta = 20$ in our experiments). After that we renormalize the probability distribution since after setting some probability bins to zero, the result is no longer a probability distribution. After we compute the new truncated probability distribution, we simply compute the expected probability bin:

\begin{equation*}
    \hat x = \frac{H}{n}\sum_{j=1}^{n}\left(j \cdot \frac{p_j \cdot \mathds{1}[|j - \arg\max p| \leqslant \Delta]}{\sum_{k=1}^{n} p_k \cdot \mathds{1}[|k - \arg\max p| \leqslant \Delta]} \right)
\end{equation*}

The above equation defines the continuous coordinate prediction $\hat{x}$ for the $x$ coordinate (valued in the range $[0, H]$). We use the same formula to compute the prediction for the $y$ coordinate denoted $\hat{y}$.

This parameterization can be very easily implemented in Jax:

\vspace{5pt}
\noindent \begin{minipage}{\linewidth}
\begin{verbatim}
import jax.numpy as jnp
def trunc_softargmax(p, delta=20):
  n = p.shape[0] # p is vector
  js = jnp.arange(n)
  j = jnp.argmax(p)
  m = jnp.abs(js - j) <= delta
  p *= m
  p /= p.sum()
  return (p * js).sum()
\end{verbatim}
\end{minipage}
\vspace{5pt}

As we mention earlier, we use two loss functions for coordinate prediction: Huber loss on the continuous prediction ($\hat x$ and $\hat y$) and softmax cross entropy for discrete prediction ($p_x$ and $p_y$):

\begin{align*}
    L(x, y, p_x, p_y) &= w_1 L_H(x, \hat x) + w_2 L_H(y, \hat y) \\
    &+ w_3 L_C(\text{one\_hot}(x), p_x) \\
    &+ w_4 L_C(\text{one\_hot}(y), p_y)
\end{align*}

$L_H$ is the huber loss, $L_C$ is the softmax with crossentropy loss. Coefficients $w_1, w_2, w_3, w_4$ are weights applied to each loss component. In our experiments, we use $w_1 = w_2 = 0.1$ and $w_3 = w_4 = 1.0$.
This combined loss is applied to every layer. This means that the $T\times Q$ point track tokens after every ViT block are fed to the coordinate heads (shared MLPs with softmax output) and the the aforementioned loss is applied to the outputs of the coordinate heads. Similarly, the visibility head which is also an MLP with the sigmoid as the final activation. The loss for this head is binary cross entropy. Like the coordinate head, the visibility head is applied to every layer. 

Note that, despite using a parameterization that produces bounded ranges for coordinate values ($(x, y) \in [0, H]\times [0, W]$), it is still possible to predict (of course, not with the same trained model) the coodinates out of the view, similarly to what other models (e.g. CoTracker \citep{cotracker}) do. For that, we could simply map e.g. the $x$ coordinate to some range $[-d, H+d]$ instead of $[0, H]$ for some positive $d$. This way, some probability bins would be responsible for out-of-view prediction while some other will do in-view prediction. Since out-of-view prediction is not a part of the TAP-Vid benchmark, we do not use it. 




\section{TAPNext Hyperparameters}
\label{app:hypers}

We sample batches containing query points of shape $[B,Q,3]$, where $B=256$ is the batch size, $Q=256$ number of query points per video. The last dimension represents the $t,x,y$ coordinate of each query point. 
Importantly, we train our model on a mixture that simulates query point being in the beginning of the video ($t=0$) and at the intermediate timestep ($t>0$). The weights of this mixture are $[0.8, 0.2]$, respectively. The $t=0$ mixture component contains query points queried at the $0^{th}$ frame in the video. The $t>0$ mixture component contains the ones queried at any timestep, not necessarily starting at the $0^{th}$ frame. Since our model is causal, it cannot track points before point query is given. Therefore, for the second component we mask the coordinate losses for frames preceding the known query and set the visibility label to zero.
We use a weight of $1.0$ for both visibility and coordinate classification losses, and $0.1$ for coordinate regression. We use the AdamW optimizer with weight decay of $0.01$ for $300,000$ steps using a cosine learning rate schedule with $2500$ warm-up steps, peak and final learning rate values of $0.001$ (for -S model) and $0$, respectively. We clip gradient norm to $1.0$. We found it important to use $8\times 8$ patches, smaller than the $16 \times 16$ used in ViT for classification, aligning with the intuition that smaller patches work well for spatially fine-grained tasks. 
\begin{table}[h!]
    \centering
    \resizebox{\linewidth}{!}{
    \begin{tabular}{c|l|c|c}
    \toprule
    & Name & TAPNext-S & TAPNext-B  \\
    \midrule
    & layers & \multicolumn{2}{c}{12} \\
    & parameters & 56M & 194M \\
    \midrule
     & attention heads & 12 & 12 \\
    \textbf{ViT Block} & width & 384 & 768 \\
    & dropout & $0.0$ & $0.0$ \\
    \midrule
     & width & 384 & 768 \\
    \textbf{SSM Block} & LRU width & 768 & 768 \\
    & heads & 12 & 12 \\
    \bottomrule
    \end{tabular}
    }
    \caption{TAPNext hyperparameters specific to each size of the model.}
    \label{tab:hps_sizes}
\end{table}

The full list of hyperparameters is available in Tables \ref{tab:hps_sizes} and \ref{tab:hps}. 
We implement the model in Jax and use TPUv6e for training. Specifically, we use $16\times 16$ TPU slice to train both TAPNext-S and TAPNext-B. Since we are using a large batch ($B\times T = 256 \times 48 = 12288$ images in our case), we use activation checkpointing (implemented in the same way as in the original ViT\footnote{\href{https://github.com/google-research/big\_vision/blob/main/big\_vision/models/vit.py}{https://github.com/google-research/big\_vision/blob/main/big\_vision/models/vit.py}}). This setup is roughly equivalent to 50 H100 GPUs in both compute and memory and our training takes 4 days for TAPNext-S and 5 days for TAPNext-B. Note that despite a large compute and memory requirement for training, TAPNext inference runs quickly on a single GPU (see Table \ref{tab:speed}).

\begin{table}[h!]
    \centering
    \resizebox{\linewidth}{!}{
    \begin{tabular}{l|c}
    \toprule
    Name & Value  \\
    \midrule
    \multicolumn{2}{c}{\textbf{Optimization}} \\
    \midrule
    Optimizer & AdamW  \\
    Global batch size & $256$  \\
    Number of queries per video & $256$ \\
    Max gradient norm & $1.0$  \\
    Weight decay & $0.01$ \\
    Number of optimization steps & $300,000$ \\
    Warmup & linear \\
    Number of warmup steps & $2500$ \\
    LR before warmup & $0$ \\
    LR schedule & cosine \\
    Peak LR (TAPNext-S) & $0.001$ \\
    Peak LR (TAPNext-B) & $0.0005$ \\
    Final LR & $0$ \\
    Precision & \texttt{float32} \\
    Regression loss weight(s) & $0.1$ \\
    Classification loss weight(s) & $1.0$ \\
    \midrule
    \multicolumn{2}{c}{\textbf{Data}} \\
    \midrule
    Dataset size (videos) & $500.000$ \\
    Dataset resolution & $256\times 256$ \\
    Number of frames per video & $48$ \\
    Camera panning & Enabled \\
    Motion blur & Enabled \\
    \makecell[l]{\small{Prob. of sampling query with $t=0$}} & 0.8 \\
    \makecell[l]{\small {Prob. of sampling query with $t>0$}} & 0.2 \\
    \midrule 
    \multicolumn{2}{c}{\textbf{Model}} \\
    \midrule 
    Patch size & $8\times 8$ \\
    Image position embedding & learned \\
    Point position embedding & sincos2d \\
    Point position embedding resolution & $256\times 256$ \\
    \makecell[l]{MLP head number of layers} & 3 \\
    \makecell[l]{MLP head hidden size} & 256 \\
    \makecell[l]{MLP head activation} & GELU \\
    Softargmax threshold ($\Delta$) & $20$ \\
    Coord. softmax temperature & $2$ \\
    Huber loss weight & $0.1$ \\
    Coordinate CE loss weight & $1.0$ \\
    Visibility CE loss weight & $1.0$ \\
    \bottomrule
    \end{tabular}
    }
    \caption{TAPNext hyperparameters}
    \label{tab:hps}
\end{table}

\section{Strided Evaluation with TAPNext as Causal Tracker}
\label{app:details_strided}
Recall that in training when TAPNext is trained on queries at $t>0$ (i.e. after the $0^{th}$ frame), the coordinate losses corresponding to frames preceding the query is masked and visible target is set to $0.0$. For complete and comparable evaluation, however, we require track predictions for every frame in the sequence. To obtain these predictions with our causal, per-frame tracker, we run the tracker both forwards and backwards in time starting at the frame corresponding to the query. These two sequences of predictions corresponding to normal and reverse time processing are then concatenated together to obtain the full sequence of prediction. For strided evaluation this process is repeated for every query point in the sequence.  

\section{Generation Pipeline for a Large Scale Synthetic Dataset}
\label{app:kubric}

\begin{figure}[t]
\centering
\setlength{\tabcolsep}{1pt}
\begin{subfigure}{\linewidth}
    \centering
    \begin{tabular}{ccc}
        \includegraphics[width=0.3\linewidth]{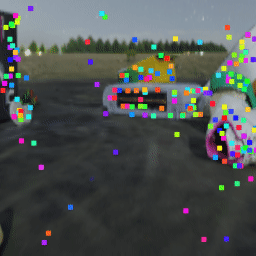} &
        \includegraphics[width=0.3\linewidth]{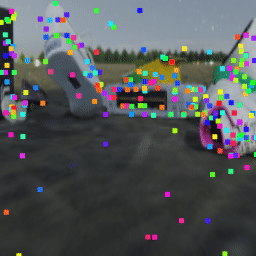} &
        \includegraphics[width=0.3\linewidth]{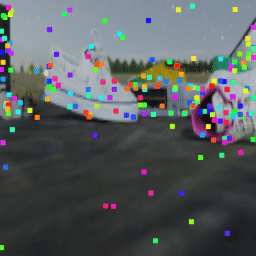} \\
        \includegraphics[width=0.3\linewidth]{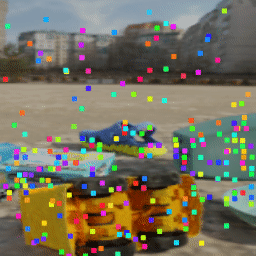} &
        \includegraphics[width=0.3\linewidth]{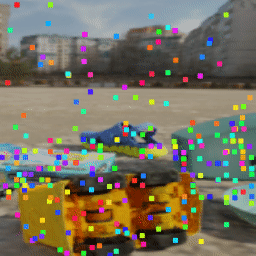} &
        \includegraphics[width=0.3\linewidth]{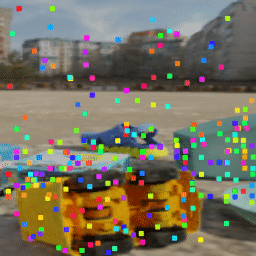}
    \end{tabular}
    \vspace{-2mm}
    \caption{Kubric MOVi-E dataset}
    \label{fig:subfig1}
\end{subfigure}

\begin{subfigure}{\linewidth}
    \centering
    \begin{tabular}{ccc}
        \includegraphics[width=0.3\linewidth]{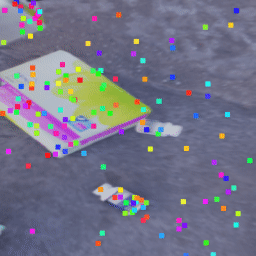} &
        \includegraphics[width=0.3\linewidth]{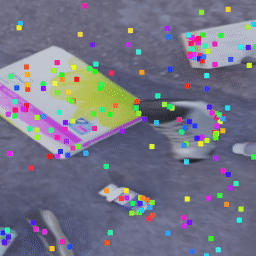} &
        \includegraphics[width=0.3\linewidth]{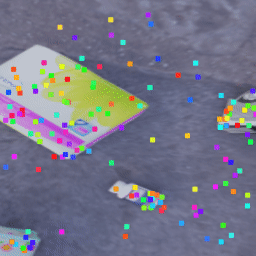} \\
        \includegraphics[width=0.3\linewidth]{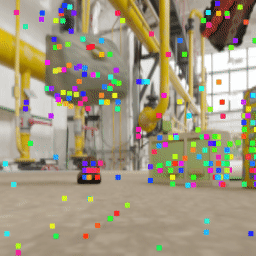} &
        \includegraphics[width=0.3\linewidth]{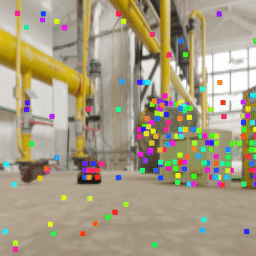} &
        \includegraphics[width=0.3\linewidth]{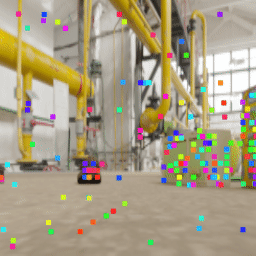}
    \end{tabular}
    \vspace{-2mm}
    \caption{Kubric MOVi-F dataset}
    \label{fig:subfig3}
\end{subfigure}

\begin{subfigure}{\linewidth}
    \centering
    \begin{tabular}{ccc}
        \includegraphics[width=0.3\linewidth]{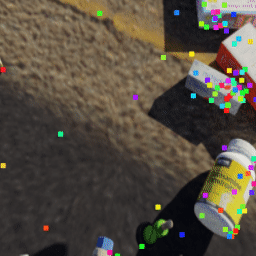} &
        \includegraphics[width=0.3\linewidth]{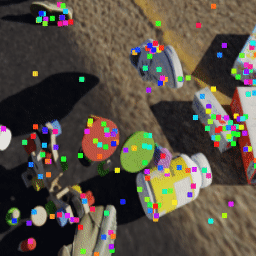} &
        \includegraphics[width=0.3\linewidth]{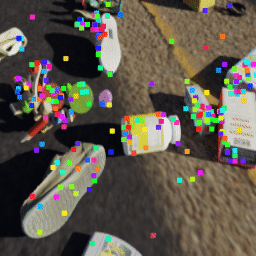} \\
        \includegraphics[width=0.3\linewidth]{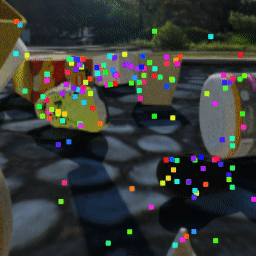} &
        \includegraphics[width=0.3\linewidth]{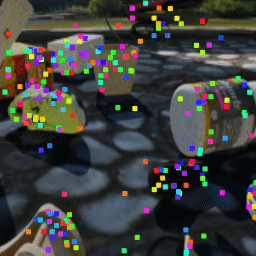} &
        \includegraphics[width=0.3\linewidth]{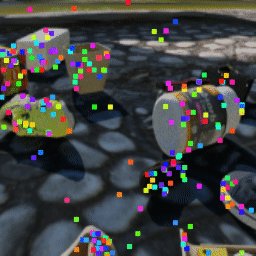}
    \end{tabular}
    \vspace{-2mm}
    \caption{MOVi-E Panning dataset}
    \label{fig:subfig2}
\end{subfigure}

\begin{subfigure}{\linewidth}
    \centering
    \begin{tabular}{ccc}
        \includegraphics[width=0.3\linewidth]{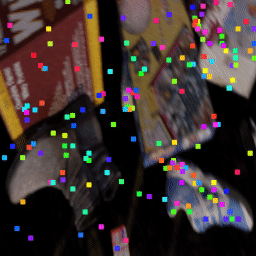} &
        \includegraphics[width=0.3\linewidth]{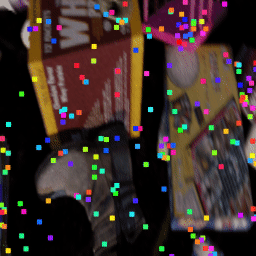} &
        \includegraphics[width=0.3\linewidth]{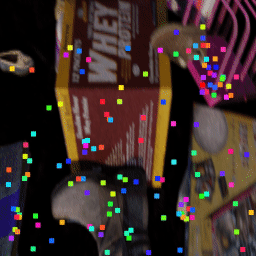} \\
        \includegraphics[width=0.3\linewidth]{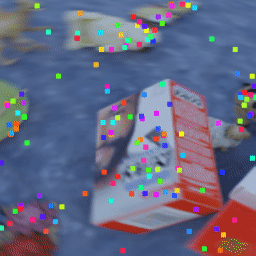} &
        \includegraphics[width=0.3\linewidth]{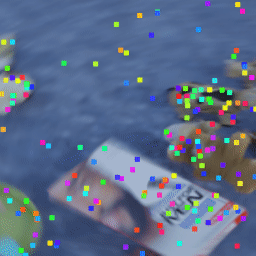} &
        \includegraphics[width=0.3\linewidth]{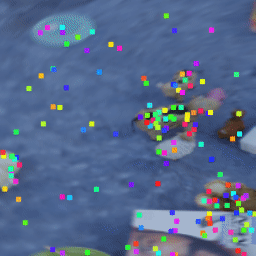}
    \end{tabular}
    \vspace{-2mm}
    \caption{MOVi-F Panning dataset (Ours)}
    \label{fig:subfig4}
\end{subfigure}
\vspace{-2mm}
\caption{
\textbf{Kubric dataset comparison} -- We show two exampler videos for each dataset with first frame, middle frame and last frame, with groundtruth point tracks. 
}
\label{fig:app:panning_movi_f}
\end{figure}

The original MOVi-F is similar to MOVi-E but includes random motion blur and is rendered at $512\times512$ resolution, with a $256\times256$ downscaled option. To enhance model performance on real-world videos with panning, we modify the MOVi-E dataset to adjust the camera’s “look at” point to follow a random linear trajectory by sampling a start point $a$ within a medium-sized sphere (4 unit radius), a travel-through point $b$ near the center of the workspace (1 unit radius), and an end point $c$ extrapolated along the line from $a$ to $b$ by up to 50\% of their distance. The "look at" path randomly switches between $a \to c$ and $c \to a$. This modification initially resulted in a 100K video Panning Kubric MOVi-E dataset. To further exploit the scalability of TAPNext and improve its generalization capability to longer real world videos, we developed a new Panning Kubric MOVi-F data generation pipeline, combining data generation pipelines from both Kubric ~\cite{greff2022kubric} and TAPIR ~\cite{tapir}. Building on this, we created a new large-scale Panning Kubric MOVi-F dataset with 500,000 videos. The rendered videos combine both panning effect and motion blur. We then increase each video duration from the default 24 frames to 48 frames. These enhancements allow more robust training and long term inference stability for TAPNext, addressing the challenges posed by real-world long term point tracking. Figure \ref{fig:app:panning_movi_f} shows the comparison between the new dataset and existing ones.

\section{Prediction Visualization on DAVIS}
\label{app:tail_visualization}

We present additional examples of point track predictions from TAPNext on the TAPVid-DAVIS dataset in Figure ~\ref{fig:app:tail_visualization}. All DAVIS track videos can be found in the supplementary files. The query points are all initialized in the first frame, and TAPNext performs tracking in a causal manner. While the model occasionally makes errors over longer time spans due to the inherent challenges of maintaining long-term precision in causal tracking, TAPNext demonstrates robust performance by reliably tracks points across a wide range of scenarios, including foreground and background elements, as well as small and large motions.

\begin{figure}[t]
\centering
\makebox[\linewidth]{
    \setlength{\tabcolsep}{1pt}
    \begin{tabular}{ccc}
        \includegraphics[width=0.33\linewidth]{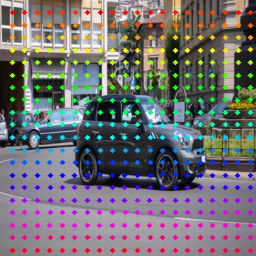} &
        \includegraphics[width=0.33\linewidth]{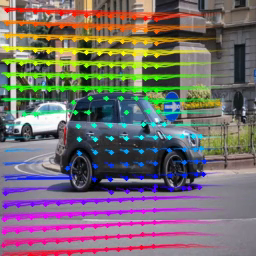} &
        \includegraphics[width=0.33\linewidth]{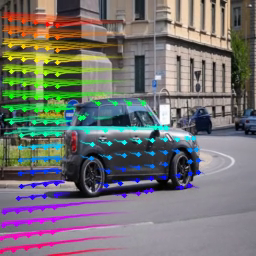} \\
        \includegraphics[width=0.33\linewidth]{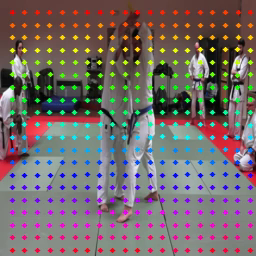} &
        \includegraphics[width=0.33\linewidth]{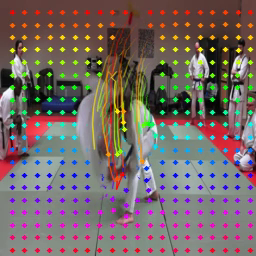} &
        \includegraphics[width=0.33\linewidth]{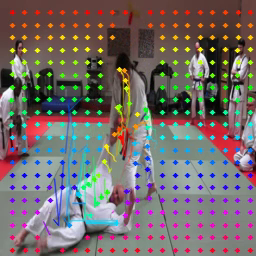} \\
        \includegraphics[width=0.33\linewidth]{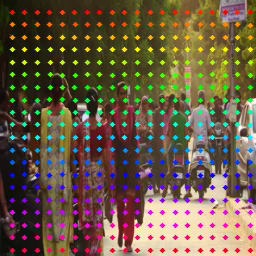} &
        \includegraphics[width=0.33\linewidth]{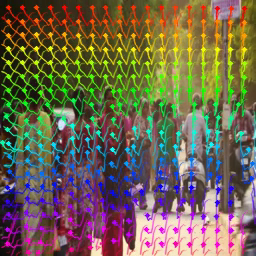} &
        \includegraphics[width=0.33\linewidth]{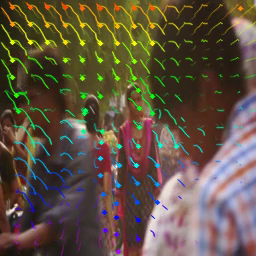} \\
        \includegraphics[width=0.33\linewidth]{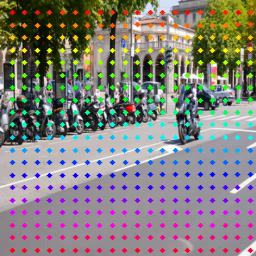} &
        \includegraphics[width=0.33\linewidth]{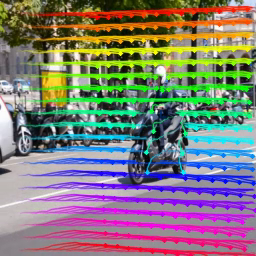} &
        \includegraphics[width=0.33\linewidth]{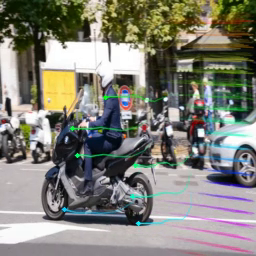} \\
        \includegraphics[width=0.33\linewidth]{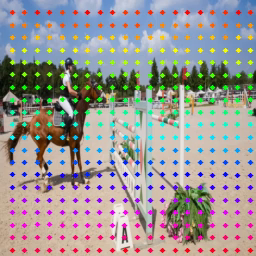} &
        \includegraphics[width=0.33\linewidth]{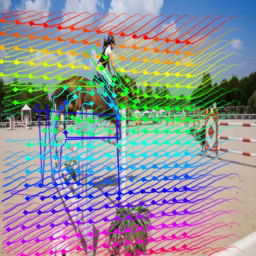} &
        \includegraphics[width=0.33\linewidth]{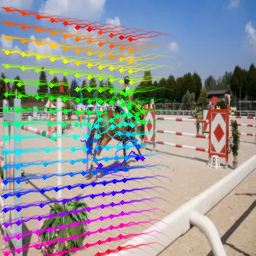} \\
    \end{tabular}
}
\caption{
\textbf{Prediction on TAPVid-DAVIS} --
We show the tail visualization of semi-dense point track predictions for 5 example videos on DAVIS dataset. For each video, we show the first query frame, 20th frame and 40th frame.
}
\label{fig:app:tail_visualization}
\end{figure}

\end{document}